\documentclass[pdflatex,sn-mathphys-num]{sn-jnl}

\usepackage{graphicx}%
\usepackage{multirow}%
\usepackage{amsmath,amssymb,amsfonts}%
\usepackage{amsthm}%
\usepackage{mathrsfs}%
\usepackage[title]{appendix}%
\usepackage{xcolor}%
\usepackage{textcomp}%
\usepackage{manyfoot}%
\usepackage{booktabs}%
\usepackage{algorithm}%
\usepackage{algorithmicx}%
\usepackage{algpseudocode}%
\usepackage{listings}%
\usepackage{array}
\usepackage{stfloats}
\usepackage{url}
\usepackage{verbatim}
\usepackage{graphicx}
\usepackage{subcaption}
\usepackage{pifont}
\usepackage{bm}

\usepackage{amssymb}
\usepackage{multirow}

\usepackage{fontawesome5}
\usepackage{overpic}
\usepackage{multirow}
\usepackage{tabularx}
\usepackage{longtable}
\usepackage{booktabs}
\usepackage{lscape}
\usepackage{orcidlink}

\theoremstyle{thmstyleone}%
\theoremstyle{thmstyletwo}%

\theoremstyle{thmstylethree}%

\begin{document}

\title{Human-to-Robot Interaction: Learning from Video Demonstration for Robot Imitation}



\author[1]{\fnm{Thanh} \sur{Nguyen Canh~\orcidlink{/0000-0001-6332-1002}}}
\equalcont{These authors contributed equally to this work.}

\author[2]{\fnm{Thanh-Tuan} \sur{Tran~\orcidlink{0009-0006-4370-8326}}}
\equalcont{These authors contributed equally to this work.}

\author[1]{\fnm{Haolan} \sur{Zhang~\orcidlink{0009-0007-1742-3754}}}

\author[1]{\fnm{Ziyan} \sur{Gao~\orcidlink{0000-0001-9948-7960}}}

\author[1, 3]{\fnm{Nak Young} \sur{Chong~\orcidlink{0000-0001-5736-0769}}}

\author*[2]{\fnm{Xiem} \sur{HoangVan~\orcidlink{0000-0002-7524-6529
}}}\email{xiemhoang@vnu.edu.vn}

\affil[1]{\orgdiv{School of Information Science}, \orgname{Japan Advanced Institute of Science and Technology}, \orgaddress{\city{Nomi}, \postcode{923-1211}, \state{Ishikawa}, \country{Japan}}}

\affil[2]{\orgdiv{University of Engineering and Technology}, \orgname{Vietnam National University}, \orgaddress{\postcode{10000}, \state{Hanoi}, \country{Vietnam}}}

\affil[3]{\orgdiv{Department of Robotics}, \orgname{Hanyang University}, \orgaddress{Ansan}, \postcode{15588}, \state{Gyeonggi}, \country{Korea}}


\abstract{Learning from Demonstration (LfD) offers a promising paradigm for robot skill acquisition. Recent approaches attempt to extract manipulation commands directly from video demonstrations, yet face two critical challenges: (1) general video captioning models prioritize global scene features over task-relevant objects, producing descriptions unsuitable for precise robotic execution, and (2) end-to-end architectures coupling visual understanding with policy learning require extensive paired datasets and struggle to generalize across objects and scenarios. To address these limitations, we propose a novel ``Human-to-Robot'' imitation learning pipeline that enables robots to acquire manipulation skills directly from unstructured video demonstrations, inspired by the human ability to learn by ``\textit{watching}'' and ``\textit{imitating}''. Our key innovation is a modular framework that decouples the learning process into two distinct stages: (1) Video Understanding, which combines Temporal Shift Modules (TSM) with Vision-Language Models (VLMs) to extract actions and identify interacted objects, and (2) Robot Imitation, which employs TD3-based deep reinforcement learning to execute the demonstrated manipulations. We validated our approach in PyBullet simulation environments with a UR5e manipulator and in a real-world experiment with a UF850 manipulator across four fundamental actions: reach, pick, move, and put. For video understanding, our method achieves $89.97\%$ action classification accuracy and BLEU-4 scores of $0.351$ on standard objects and $0.265$ on novel objects, representing improvements of $76.4\%$ and $128.4\%$ over the best baseline, respectively. For robot manipulation, our framework achieves an average success rate of $87.5\%$ across all actions, with $100\%$ success on reaching tasks and up to $90\%$ on complex pick-and-place operations. These results demonstrate strong generalization capability, particularly for previously unseen object categories. The project website is available at \url{https://thanhnguyencanh.github.io/LfD4hri}.
}


\maketitle

\section{Introduction}\label{sec:introduction}

The rapidly expanding application of robotics in modern industry and domestic settings necessitates alternative, efficient methods for skill acquisition, moving beyond time-consuming procedural coding. However, robotic skill acquisition has traditionally relied on meticulous programming or complex mathematical modeling of domain dynamics~\cite {eze2025learning}. Learning from Demonstration (LfD), also referred to as Imitation Learning (IL), offers a promising alternative, enabling robots to acquire new policies simply by observing examples provided by human teachers without tedious programming~\cite{calinon2018learning, argall2009survey}. LfD frameworks typically decouple the learning process into two essential phases: gathering demonstration examples and deriving an executable policy. Early LfD approaches, such as Kinesthetic Teaching (KT)~\cite{akgun2012trajectories, sakr2020training, gomez2012kinesthetic}, involve the teacher physically guiding the robot. While KT provides dense, high-precision trajectory data, often eliminating the need for mapping between teacher and robot bodies~\cite{akgun2012trajectories, maccio2024kinesthetic}, it is inherently non-scalable, labor-intensive, and limits data to the robot's existing kinematics. Consequently, modern efforts prioritize observational learning from unstructured real-world videos to unlock scalable, versatile skill transfer, including teleoperation~\cite{fu2023teleoperation, yang2016teleoperation} and passive observation~\cite{koenemann2014real, peters2016feasibility, racinskis2022motion}. While teleoperation widely uses demonstration input for trajectory or pose learning, it typically requires external input for the robot through joysticks, handlers, or pose tracking tools; passive observation uses a bodysuit or sensor device to capture human motions,  which generally requires constant time steps to acquire sequential data (\textit{e.g.}, end-effector position, orientation, and gripper signals). Although effective, these traditional LfD methods face practical limitations, including dependence on specialized hardware (\textit{e.g.}, motion capture suits or haptic devices) and difficulty scaling to diverse tasks~\cite{yang2023watch}.

\begin{figure*}[t!]
    \centering
    \begin{overpic}[width=\textwidth, unit=1pt]{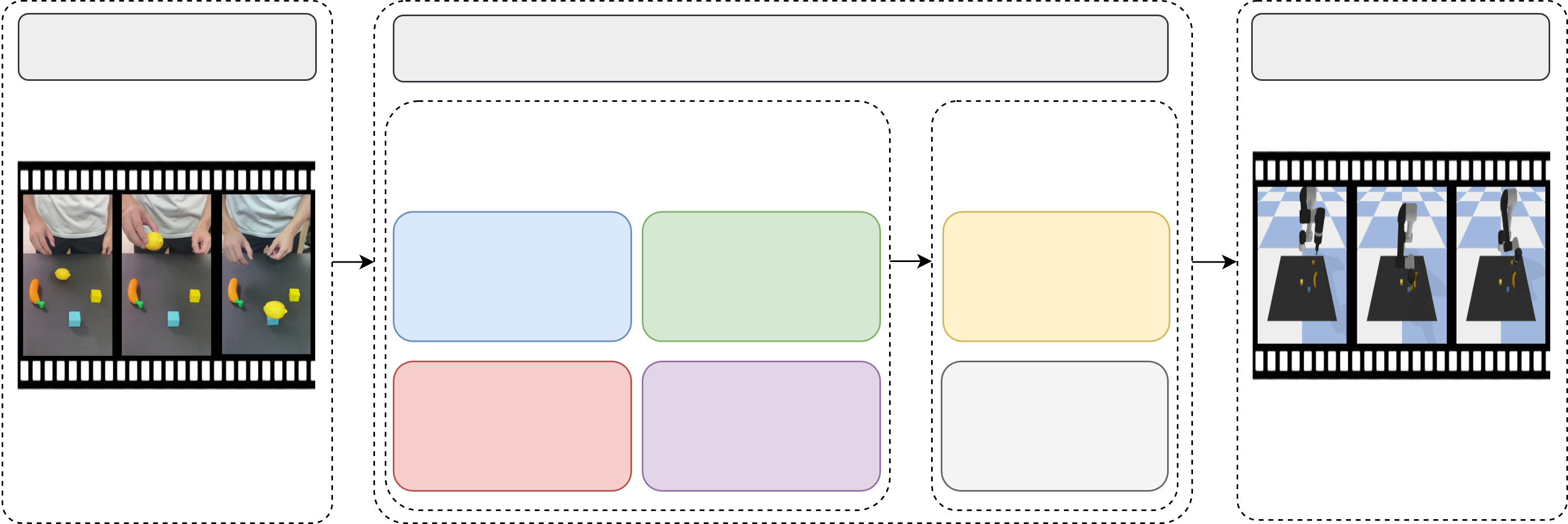}
        \put(7.4, 29.8){\footnotesize \textbf{Inputs}}
        \put(7.4, 3.0){\footnotesize Video}
        \put(32.0, 29.8){\footnotesize \textbf{Imitation Learning Architecture}}
        \put(80.4, 29.8){\footnotesize \textbf{Robot Execution}}
        \put(86.3, 3.0){\footnotesize Robot}
        \put(29.7, 23.4){\footnotesize Video Understanding}
        \put(59.9, 23.4){\footnotesize Robot Learning}
        \put(64.3, 16.5){\footnotesize Policy}
        \put(63.2, 14.0){\footnotesize Learning}
        \put(64.3, 7.0){\footnotesize Motion}
        \put(63.2, 4.5){\footnotesize Planning}
        \put(25.8, 16.5){\footnotesize Visual encoder}
        \put(27.5, 14.0){\footnotesize CNN/ViT}
        \put(29.5, 7.0){\footnotesize Action}
        \put(27.0, 4.5){\footnotesize Recognition}
        \put(41.2, 16.5){\footnotesize Pose estimation}
        \put(43.5, 14.0){\footnotesize \& Tracking}
        \put(42.3, 5.7){\footnotesize VLMs/LLMs}
    \end{overpic}
    \caption{An illustration of Robot Imitation Learning from Human Demonstration.}
    \label{fig:ILexample}
\end{figure*}
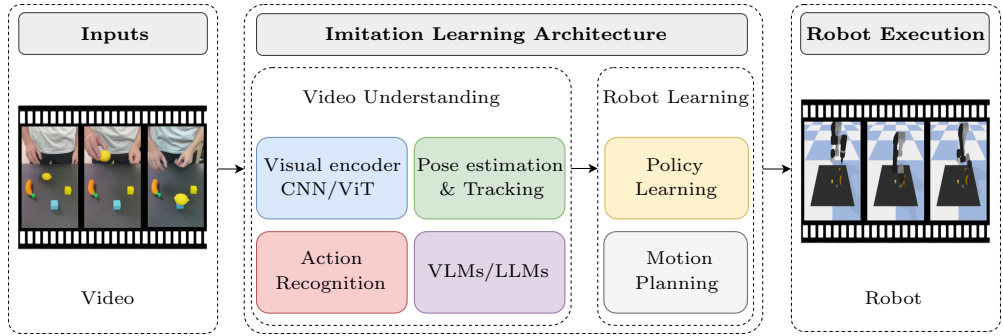

Humans, in contrast, possess the remarkable ability to learn new skills from diverse, non-egocentric visual observations, such as simply watching a video clip. Critically, humans tend to extract the goal from a demonstration first, then use mastered skills to achieve it. This suggests that decoupling demonstration understanding from skill learning may be a more effective and robust approach than attempting to map visual data directly to manipulation policies in a single, unified framework. Inspired by the human ability to learn behaviors simply by ``watching and imitating'', which is illustrated in Fig.~\ref{fig:ILexample}, recent research~\cite{yang2023watch, venugopalan2015sequence} has pivoted towards leveraging easily obtainable visual demonstrations for robot skill transfers. This approach transforms unstructured visual data into actionable instructions, often framed as the video-to-command problem~\cite{nguyen2019v2cnet}. However, video-to-command for robot manipulation introduces two distinct, difficult challenges that must be overcome for successful development: (1) Demonstration Understanding (The ``What''), and (2) Execution and Embodiment (The ``How''). In the first stage, the robot must accurately interpret the video content to generate a correct, executable command~\cite{karpathy2014large}. General video captioning approaches~\cite{nguyen2018translating, yu2018one}, which typically rely on uniformly sampled frames, often fail in complex scenarios involving changeable illumination, varied backgrounds, and diverse objects. Crucially, these general models lack the precision needed for robotic applications because they focus on global scene features rather than specific manipulated objects~\cite{yang2016teleoperation, smith2019avid}. Furthermore, early video-to-command models relying on LSTM-based architectures tend to overfit due to the repetitiveness in video descriptions, limiting generalization~\cite{, zhang2019reconstruct, yang2023watch}. Even with an accurate command, the robot must robustly translate semantic concepts like ``scoop'' or ``fold'' grounded in the human's environment into precise, sub-centimeter spatial actions appropriate for its own body - the embodiment mapping problem~\cite{shridhar2022cliport}. This process requires converting high-level text commands into reliable low-level manipulation frames or accurate affordance predictions~\cite{do2018affordancenet}, a challenging step that requires sophisticated spatial reasoning.

To bridge the gap between flexible visual demonstrations and robust execution, we propose a novel and modular ``Human-to-Robot'' imitation learning pipeline. This pipeline explicitly decouples the learning process into two specialized stages: (1) \textit{Video Understanding}---accurately interpreting the video $\mathcal{V}$ content to identify actions, objects, and interactions (the ``what''), and (2) \textit{Robot Imitation}---effectively translating this interpretation into executable manipulation policies that adapt to various scenarios (the ``how''). Unlike prior studies~\cite{nguyen2018translating, nguyen2019v2cnet, yang2023watch}, which rely heavily on extensive data preparation and end-to-end architectures, our approach decomposes complex movements into fundamental action primitives and explicitly handles scene distractions involving multiple objects. This modular structure enables robots to acquire diverse manipulation skills solely from visual demonstrations while achieving strong generalization across scenarios. The main contributions of this work are summarized as follows:

\begin{enumerate}
    \item \textbf{A Decoupled Human-to-Robot Imitation Learning Pipeline.} We propose a novel two-stage architecture that separates demonstration understanding from policy learning. This decoupling eliminates the need for paired human-robot data or explicit kinematic correspondence, enabling scalable skill transfer from unstructured videos to robotic manipulation.
    
    \item \textbf{Object-Centric Video Understanding with Temporal Modeling.} We introduce a captioning framework that generates concise, grammar-free command sentences (\textit{e.g.}, ``place apple on plate'') instead of complex natural language descriptions. The framework comprises: (a) an \textit{Action Understanding Module} integrating Temporal Shift Modules (TSM) with ResNet backbones to capture fine-grained motion dynamics across ten fundamental actions, and (b) an \textit{Interacted-Objects Understanding Module} employing a specialized Object Selection algorithm with blur detection and overlap minimization to identify target objects, followed by Vision-Language Model (VLM) integration for robust object recognition and zero-shot generalization.
    
    \item \textbf{Hierarchical Reward Design for Precise Manipulation.} We develop a TD3-based deep reinforcement learning framework with a carefully structured reward function that decomposes complex manipulation into learnable sub-goals. The hierarchical design incorporates approach rewards, interaction rewards, and alignment rewards, along with safety penalties for collisions, workspace violations, and motion smoothness, enabling sub-centimeter positioning accuracy in cluttered environments.
\end{enumerate}

The remainder of this paper is organized as follows. Section~\ref{sec:relatedworks} reviews related work in demonstration understanding, video-to-language models, and robotic affordance learning. Section~\ref{sec:proposed} details our proposed modular pipeline, including the algorithms for both Video Understanding and Robot Imitation stages. Section~\ref{sec:result} presents experimental setup, comparative evaluations, and ablation studies. Finally, Section~\ref{sec:conclusion} concludes the paper with a discussion of limitations and future research directions.

\section{Related Work}
\label{sec:relatedworks}
\subsection{Learning from Demonstration Paradigms}
LfD approaches are typically categorized by how demonstration data is collected, involving two phases: gathering examples and deriving an executable policy. Traditional methods rely on physical interfaces for collecting precise trajectories. \textit{Kinesthetic teaching}~\cite{akgun2012trajectories, sakr2020training, maccio2024kinesthetic, gomez2012kinesthetic} involves the human teacher directly guiding the robot through desired movements. While this approach provides dense, high-precision trajectory data, it is laborious and time-consuming, often requiring prior training for effective demonstrations. Recent work~\cite{maccio2024kinesthetic} explores kinesthetic teaching in Mixed Reality environments to improve robot intent communication. \textit{Teleoperation}~\cite{yang2016teleoperation, fu2023teleoperation} employs external devices such as haptic interfaces, joysticks, or pose-tracking tools for remote robot control. These methods often incorporate wave variables and neural networks to handle system complexities including time-varying delays. \textit{Passive observation} or Motion Capture (MoCap)~\cite{koenemann2014real, peters2016feasibility, racinskis2022motion} records human actions through vision systems, external sensors, or dedicated motion capture suits. Racinskis \textit{et al.}~\cite{racinskis2022motion} explore affordance-based learning from video data using Cartesian space representations, while Peters \textit{et al.}~\cite{peters2016feasibility} evaluate the feasibility of training robots purely through passive observation.

A fundamental challenge in LfD is the \textit{correspondence problem}, which addresses the mismatch between teacher and learner domains~\cite{correia2024survey}. This includes \textit{embodiment mapping}—whether states and actions from human demonstrators match those the robot would observe or execute. This challenge is particularly acute when learning from human video, as simple imitation fails due to significant visual and kinematic domain shifts~\cite{hwang2018real, liu2018imitation}. Prior work~\cite{smith2019avid} addresses this through image-to-image translation methods such as CycleGANs~\cite{zhu2017unpaired}. In contrast, our work resolves this challenge through linguistic and geometric decoupling, making visual input executable regardless of domain gaps.

\subsection{Video-to-Command and Demonstration Understanding}
Moving beyond specialized hardware, the field has increasingly adopted unstructured visual demonstrations, leading to the \textit{video-to-command} problem~\cite{nguyen2018translating, nguyen2019v2cnet, zhao2025taste, mao2025robot}. Early approaches utilized two-stage pipelines: identifying semantic content (subject, verb, object) followed by template-based sentence generation~\cite{venugopalan2015sequence}. However, template-based methods inadequately capture the complexity of natural language. Sequence-to-Sequence models~\cite{venugopalan2015sequence}, particularly LSTM-based architectures, enabled direct mapping from video features to sentences.

For robotic manipulation, previous video-to-command models demonstrated limited performance due to their focus on full-frame understanding rather than task-relevant objects and fine-grained actions~\cite{nguyen2018translating}. Nguyen \textit{et al.}~\cite{nguyen2019v2cnet} proposed an additional action classification branch operating concurrently with feature extraction. Based on this insight, Yang \textit{et al.}~\cite{yang2023watch} introduced Visual Change Maps (VCM) to isolate interacted objects by computing frame differences, leveraging Mask R-CNN~\cite{jang2017end} for object segmentation. The VCM is designed to isolate interacted objects by computing the difference between consecutive frames (\textit{i.e.}, subtracting the next frame from the current frame), thereby capturing only changes related to interactions while mitigating noise and irrelevant objects. While achieving state-of-the-art performance, this method faces challenges, including data collection difficulty, LSTM overfitting due to repetitive annotations, heavy reliance on large-scale annotated datasets, and the limited number of input frames per video.

To address these limitations, we leverage the Temporal Shift Module (TSM)~\cite{lin2019tsm}, which provides computationally efficient spatio-temporal modeling through channel shifting, achieving strong performance on fine-grained human-object interaction datasets. Furthermore, we propose a modular approach that decouples action understanding from object recognition, utilizing blur detection and overlap minimization to identify task-relevant objects.

\subsection{Deep Reinforcement Learning for Robot Execution}
The challenge of ``Robot Imitation'' -how to translate commands into action- requires effective policy learning and spatial reasoning. The execution phase depends heavily on \textit{affordance} detection, which is the understanding of an object's functional properties~\cite {zhai2021one}.  Deep learning approaches using CNNs for grasp detection~\cite{lenz2015deep, zeng2022robotic} and end-to-end affordance detection methods such as AffordanceNet~\cite{do2018affordancenet} have proven effective but require substantial human-annotated training data. However, their approach requires substantial training data preparation, which is both time-consuming and challenging due to its reliance on human-annotated datasets.

Prior affordance-detection methods predominantly assign single affordance labels to object regions, overlooking multifunctionality. For instance, a block may possess both \textit{grasp} and \textit{place} affordances simultaneously. To reduce dependency on manual annotations, reinforcement learning has emerged as a viable alternative~\cite{zeng2018learning}. Recent methods~\cite{zhang2020grasp, yang2023watch} introduce RL-based frameworks for learning grasp and place affordances, while some works~\cite{kalashnikov2018scalable, zhang2020grasp} demonstrate the effectiveness of deep RL for vision-based manipulation. However, its limited generalization capability to novel tasks renders it ineffective and unsuitable for real-world applications. 

\subsection{Integrated Vision-Language/Large-Language Architectures} 
To link semantic understanding to physical execution, recent work~\cite{zitkovich2023rt, driess2023palm, zhou2024visual, jin2024robotgpt, qu2025spatialvla} moves toward integrated Vision-Language Models (VLMs) or Large-Language Models (LLMs). Inspired by biological principles of visual processing—the ventral stream (what/semantic recognition) and the dorsal stream (where/spatial processing)—architectures have emerged.  Jang \textit{et al.} ~\cite{jang2017end} use a two-stream model in which one stream focuses on grasping (dorsal) and the other on object classification (ventral). This architecture is designed to manage the trade-off between grasping accuracy and semantic understanding, sometimes leveraging auxiliary large-scale datasets like ImageNet~\cite{deng2009imagenet} for generalization. In addition, Socratic Models~\cite{zeng2022socratic} propose integrating multiple pretrained models via language-based exchange, whereas Inner Monologue~\cite{huang2022inner} focuses on incorporating environmental feedback to support adaptive behavior and scene understanding. The contemporary CLIPORT~\cite{shridhar2022cliport} architecture successfully applies this two-stream decomposition for language-conditioned robotic tasks. It leverages the contrastive pre-training of CLIP to align vision and language features, using the element-wise product of visual and language encodings to maintain spatial dimensions for precise action grounding. It leverages a pre-trained vision–language model to provide semantic grounding that maps high-level language instructions to low-level, action-relevant representations, and uses CLIP-based contrastive pre-training to align visual and linguistic features via element-wise fusion, preserving spatial structure for precise action grounding.

Unlike these methods, we develop our system based on VLMs'~\cite {liu2023visual} ability to understand and generate object categories by removing noise and irrelevant objects. Our proposed pipeline directly builds on the success of the VLM approach by employing a similar decoupled structure—Video Understanding feeds semantic commands to the Robot Imitation stage, which employs a DRL architecture for robust spatial and semantic control, further enhancing generalization capability beyond single-stream or general video-to-command systems.

\section{Proposed Method}
\subsection{System Architecture}

\label{sec:proposed}
The proposed architecture is a modular ``Human-to-Robot'' pipeline, which explicitly designed to enable robust skill acquisition from visual demonstration by separating the complexity into two specialized stages: (1) Video Understanding - determining what the human is doing, and (2) Robot Imitation - determining how the robot should execute the action.  This decoupled approach is intended to overcome the heavy reliance on extensive data preparation seen in prior studies and enhance generalization across diverse scenarios. We formally define the overall task by decomposing each step into two subtasks: understanding the demonstration video (solved by Action Understanding, Interacted-Objects Understanding, and VLM), and learning the demonstrated manipulation via DRL. 

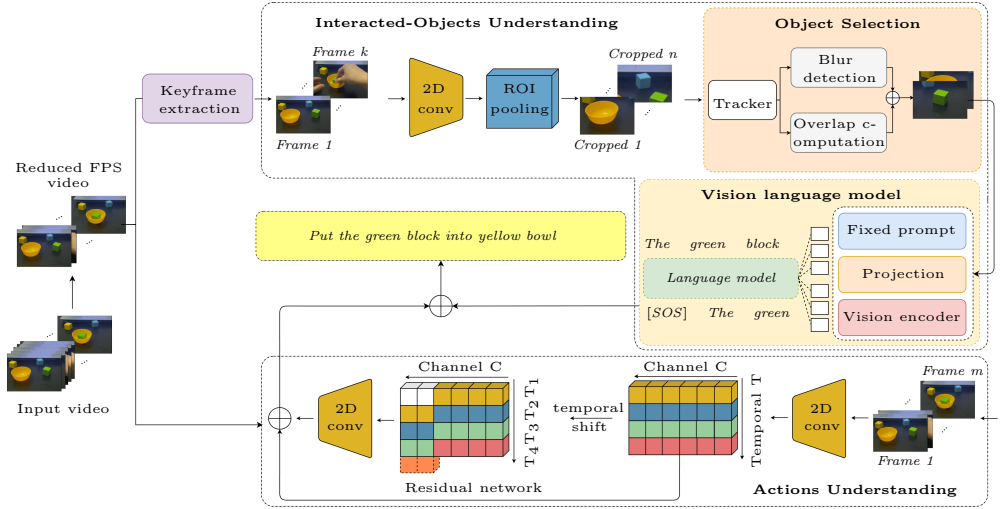
\begin{figure*} [!ht]
\centering
    \begin{overpic}[width=\textwidth, unit=1pt]{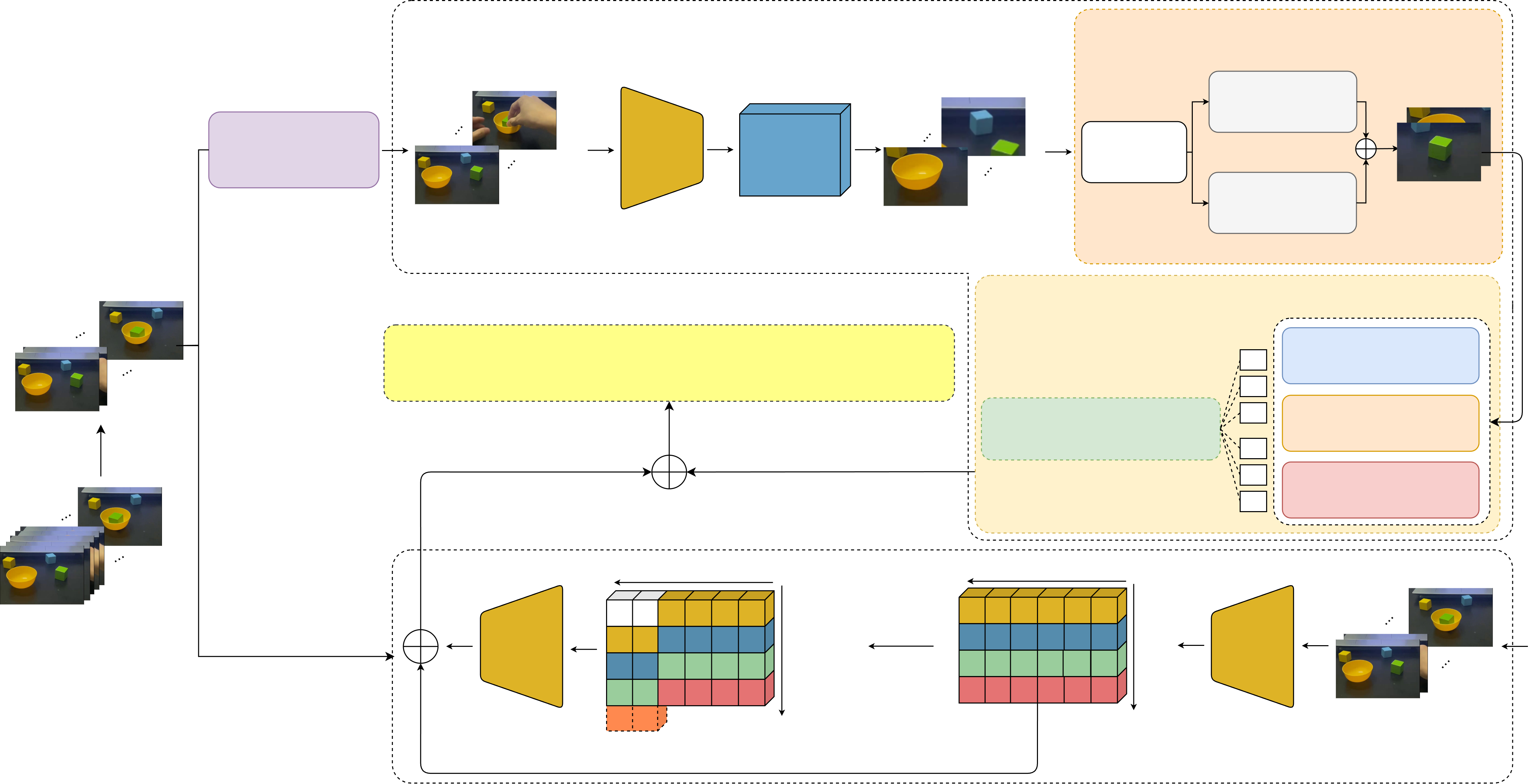}
        \put(1.0, 9.5){\tiny Input video}
        \put(0.8, 34.0){\tiny Reduced FPS}
        \put(4.0, 32.5){\tiny video}
        \put(15.4, 41.8){\tiny Keyframe }
        \put(15.3, 40.0){\tiny extraction}
        \put(31.0, 48.7){\tiny \textbf{Interacted-Objects Understanding}}
        \put(27.0, 36.3){\tiny \textit{Frame 1}}
        \put(30.6, 45.7){\tiny \textit{Frame k}}
        \put(41.8, 42.0){\tiny 2D}
        \put(41.3, 40.0){\tiny conv}

        \put(49.9, 42.0){\tiny ROI}
        \put(48.6, 40.0){\tiny pooling}

        \put(56.9, 36.3){\tiny \textit{Cropped 1}}
        \put(60.4, 45.5){\tiny \textit{Cropped n}}

        \put(77.2, 48.6){\tiny \textbf{Object Selection}}
        \put(71.0, 40.5){\tiny Tracker}
        \put(81.6, 44.9){\tiny Blur}
        \put(80.1, 43.1){\tiny detection}
        \put(79.1, 38.4){\tiny Overlap c-}
        \put(79.1, 36.6){\tiny omputation}

        \put(69.8, 31.2){\tiny \textbf{Vision language model}}
        \put(84.5, 27.7){\tiny Fixed prompt}
        \put(86.0, 23.2){\tiny Projection}
        \put(84.1, 18.9){\tiny Vision encoder}
        \put(64.0, 26.3){\tiny \textit{The} \ \ \textit{green} \ \ \textit{block}}
        \put(66.3, 22.8){\tiny \textit{Language model}}
        \put(64.3, 19.3){\tiny [\textit{SOS}] \ \ \textit{The} \ \ \textit{green}}

        \put(30.3, 27.1){\tiny \textit{Put the green block into yellow bowl}}

        \put(75.0, 1.3){\tiny \textbf{Actions Understanding}}
        \put(87.5, 4.3){\tiny \textit{Frame 1}}
        \put(92.0, 13.4){\tiny \textit{Frame m}}
        \put(80.8, 9.5){\tiny 2D}
        \put(80.2, 7.8){\tiny conv}
        \put(33.0, 9.5){\tiny 2D}
        \put(32.2, 7.8){\tiny conv}
        \put(41.3, 13.8){\tiny Channel C}
        \put(64.0, 13.8){\tiny Channel C}
        \put(74.9, 4.0){\tiny \rotatebox{90}{Temporal T}}
        \put(54.9, 9.6){\tiny temporal}
        \put(56.8, 7.7){\tiny shift}
        \put(52.1, 4.0){\tiny \rotatebox{90}{T$_4$T$_3$T$_2$T$_1$}}

        \put(40.0, 1.5){\tiny Residual network}
    \end{overpic}
\caption{The proposed Video Understanding architecture consists of two parallel branches: (a) Interacted Object Understanding Module and (b) Action Understanding Module. Initially, the raw input frames $\mathbf{F} = \{\mathbf{f}_1, \mathbf{f}_2, \dots, \mathbf{f}_n\}$ are downsampled to $\tilde{\mathbf{F}} = \{\tilde{\mathbf{f}}_1, \tilde{\mathbf{f}}_2, \dots, \tilde{\mathbf{f}}_m\}$ (where $n>m$) to optimize runtime and match the training data frame rate in Action Understanding Module.
(a) The Interacted Object Understanding Module processes $\tilde{\mathbf{F}}$ to extract a subset of keyframes $\hat{\mathbf{F}} = \{\hat{\mathbf{f}}_1, \hat{\mathbf{f}}_2, \dots, \hat{\mathbf{f}}_k\}$, (where $m>k$). These frames are analyzed by our Object Selection algorithm and Vision-Language Models (VLMs) to identify the specific objects involved in the interaction accurately.
(b) The Action Understanding Module is implemented based on a CNN architecture with a ResNet-50~\cite{he2016deep} backbone with Temporal Shift Modules (TSM) that shift feature channels along the temporal dimension to capture fine-grained motion dynamics for action classification.
}
\label{fig:videounderstanding}
\end{figure*}

Firstly, the task of demonstration understanding is formulated as a video-to-command translation problem, where the objective is to map an unstructured demonstration video $\mathcal{V}$ to a structured execution command. As illustrated in Fig.~\ref{fig:videounderstanding}, our framework decomposes this task into two parallel branches: Actions ($\mathcal{A}_i)$ Understanding and Interacted-Objects ($\mathcal{I}_i$) Understanding. The first branch, Action Understanding Module, processes the visual data to identify the manipulation primitive. Let the raw input video be presented by a sequence of frames $\mathbf{F} = \{\mathbf{f}_1, \mathbf{f}_2, \dots, \mathbf{f}_n\}$. To optimize computational efficiency, we first downsample this sequence to a reduced set $\tilde{\mathbf{F}} = \{\tilde{\mathbf{f}}_1, \tilde{\mathbf{f}}_2, \dots, \tilde{\mathbf{f}}_m\}$ (where $m <n$). From $\tilde{\mathbf{F}}$, we extract a sequence of feature vectors $\mathbf{X} = \{\mathbf{x}_1, \mathbf{x}_2, \dots, \mathbf{x}_m\}$, where each $\mathbf{x}_i$ corresponds to the feature representation of the $i^{\text{th}}$ frame. To capture fine-grained motion dynamics and temporal dependencies, we introduce a shifted feature sequence $\tilde{\mathbf{X}} = \{\tilde{\mathbf{x}}_1, \tilde{\mathbf{x}}_2, \dots, \tilde{\mathbf{x}}_m\}$, where $\tilde{\mathbf{x}}_i$ represents the temporally shifted feature of the $i^{\text{th}}$ frame. This is achieved by applying random channel shifts along the temporal dimension. This mechanism enables the model to effectively encode the human-object interactions required to distinguish between the ten fundamental manipulation actions: \textit{pick}, \textit{move}, and \textit{put}.

The second branch, Interacted-Objects Understanding module, focuses on isolating the objects involved in the interaction. Before passing frames to the Human-Object Interaction (HOI) algorithm~\cite{shan2020understanding}, the reduced sequence $\tilde{F}$ undergoes a keyframe extraction process defined as:
\begin{equation}
\tilde{\mathbf{F}}_k = T_G(\mathbf{f}_s(\mathbf{f}_t(\tilde{\mathbf{F}}))),
\end{equation}
where $T_G$ represents grayscale transformation, $\mathbf{f}_s$ denotes a subtraction function applied between consecutive frames (assuming that the change in brightness does not substantially impact the process (\textit{i.e.}, $\tilde{\mathbf{f}}_i - \tilde{\mathbf{f}}_{i-1}$), and $\mathbf{f}_t$ is a threshold filter applied to retain only significant visual changes. The processed keyframes are subsequently fed into an Human-Object Interaction (HOI) algorithm to detect objects interacting with human hands. The detected objects are then ranked based on their relevance and interaction characteristics.

The final output is a sequence of words $\mathcal{S} = \{s_1, s_2, \dots, s_m\}$, representing the generated command. We integrate the outputs from both branches to convert the video into an actionable instruction. Following standard video captioning approaches~\cite{venugopalan2015sequence}, the optimal command for a given video $\mathcal{V}$ is determined by maximizing the log-likelihood of the action-object pair $\mathcal{P}$, conditioned on the video and model parameters $\theta$:
\begin{equation}
    \theta = \arg\max_\theta \sum_{(\mathcal{V,P})} \log p(\mathcal{P} \mid \mathcal{V}, \theta).
\end{equation}
To ensure the system is robust for robotic control, we enforce a grammar-free output format rather than natural language sentences. Instead of generating complex descriptions like ``A man is picking the blue block up on the table,'' our system produces concise, execution-focused commands such as ``Picking the blue block up''. This format explicitly emphasizes the essential components—the action and the target object—ensuring clarity and direct applicability for downstream robotic manipulation tasks.

Secondly, humans naturally abstract manipulation commands such as \textit{``get a scoop of coffee''} or visual demonstrations like \textit{``fold the cloth''} by leveraging intuitive understanding of object properties and spatial relationships. In contrast, human-like robotic systems lack this innate grounding capability and must explicitly learn geometric representations, spatial awareness, and motion dynamics to perform manipulation tasks correctly. To address this challenge, we formulate robot control directly in joint space rather than Cartesian space. This design choice offers several advantages: (1) it captures the underlying motion dynamics more naturally, (2) it avoids kinematic singularities inherent in end-effector control, and (3) it reduces the gap between semantic commands and low-level motor control. For tasks requiring explicit geometric reasoning, model-based approaches can complement our learned policy. We employ a Deep Reinforcement Learning (DRL) strategy based on Twin Delayed Deep Deterministic Policy Gradients (TD3) \cite{fujimoto2018addressing}, where the objective is to maximize the accumulated discounted reward:
\begin{equation}
J(\pi_\theta) = \mathbb{E} \left[ \sum_{t=0}^\infty \gamma^t r(\mathbf{s}_t, \mathbf{a}_t) \right],
\label{eq:objective_function}
\end{equation}

We chose TD3 over alternative methods for three key reasons. First, TD3 addresses the overestimation bias prevalent in Q-learning methods by employing a twin-critic architecture, resulting in more stable value estimates. Second, its deterministic policy formulation is well-suited for continuous control tasks with high-dimensional action spaces fully randomized environment. Third, the delayed policy updates reduce variance during training, which is particularly important given our extensive domain randomization strategy.

\subsection{Video Understanding}

\subsubsection{Interacted-Objects Understanding Module (IOUM)}
The Interacted-Objects Understanding Module (IOUM) constitutes the one specialized branch of our Video Understanding framework, running in parallel with the Action Understanding Module. Its fundamental purpose is to accurately identify and classify the functional roles of the objects involved in the human manipulation demonstration, providing the interacted-object category vectors ($\mathcal{I}_j$) necessary for command generation. This object-centric focus is crucial because general captioning models often prioritize global scene features, lacking the precision needed for robotic execution. The filtered keyframes ($\hat{\mathbf{F}}$) are first passed through a CNN backbone to extract dense, high-dimensional visual feature maps. This process involves a series of 2D conv that learn hierarchical representations from the visual data. This initial convolutional step transforms the raw pixel data into a spatial feature map suitable for detecting regions of interest. Following the initial feature extraction, the network must localize objects and extract fixed-size feature vectors from their bounding boxes (Regions of Interest, RoIs). The bounding boxes are typically determined by an object detection or Human-Object Interaction algorithm. These features are standardized using ROI pooling, which extracts and pools the features corresponding to each detected bounding box from the feature map into a smaller, standardized representation. This step is critical because it ensures that features for objects of varying scales and positions are represented consistently, preparing them for subsequent classification and selection tasks. The output of this pooling step results in the feature set $\mathbf{I}= \{{\mathbf{i}}_1, {\mathbf{i}}_2, \dots, {\mathbf{i}}_h\}$, where $h$ represents the number of high-confidence objects, eliminates surrounding noise, and contains high-confidence object features.
\begin{figure*}[!ht]
    \centering
    \begin{overpic}[width=\textwidth, unit=1pt]{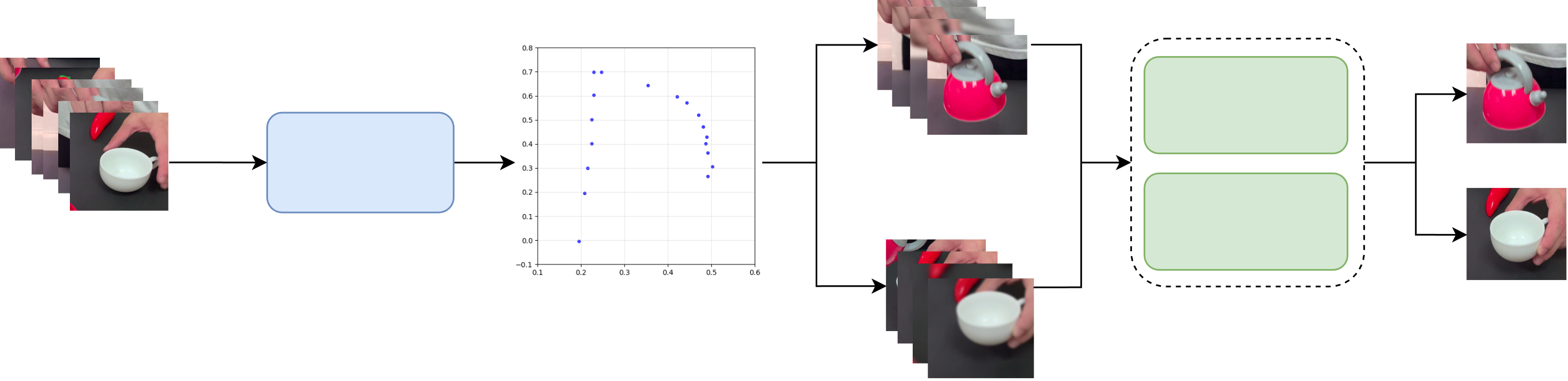}
        \put(-0.5, 9.0){\tiny Cropped object set}
        \put(20.0, 13.5){\tiny Tracker}
        \put(37.0, 4.0){\tiny Trajectory}
        \put(52.7, 25.0){\tiny Pickable-object set}
        \put(52.5, 10.0){\tiny Placeable-object set}
        \put(73.8, 17.0){\tiny Blur detection}
        \put(76.0, 11.0){\tiny Overlap}
        \put(75.5, 9.0){\tiny detection}
        \put(88.0, 4.5){\tiny Acceptable objects}
        
    \end{overpic}
    \caption{ Pipeline of the proposed Object Selection algorithm. Cropped object regions are tracked across frames to generate trajectory data, which is used to classify objects into Pickable-object and Placeable-object sets based on motion patterns. Each candidate object then undergoes Blur Detection and Overlap Detection to select the highest-quality object instances for subsequent Vision-Language Model processing.}
    \label{fig:object_selection}
\end{figure*}

\begin{figure*}[!ht]
    \centering
    \includegraphics[width=0.75\textwidth]{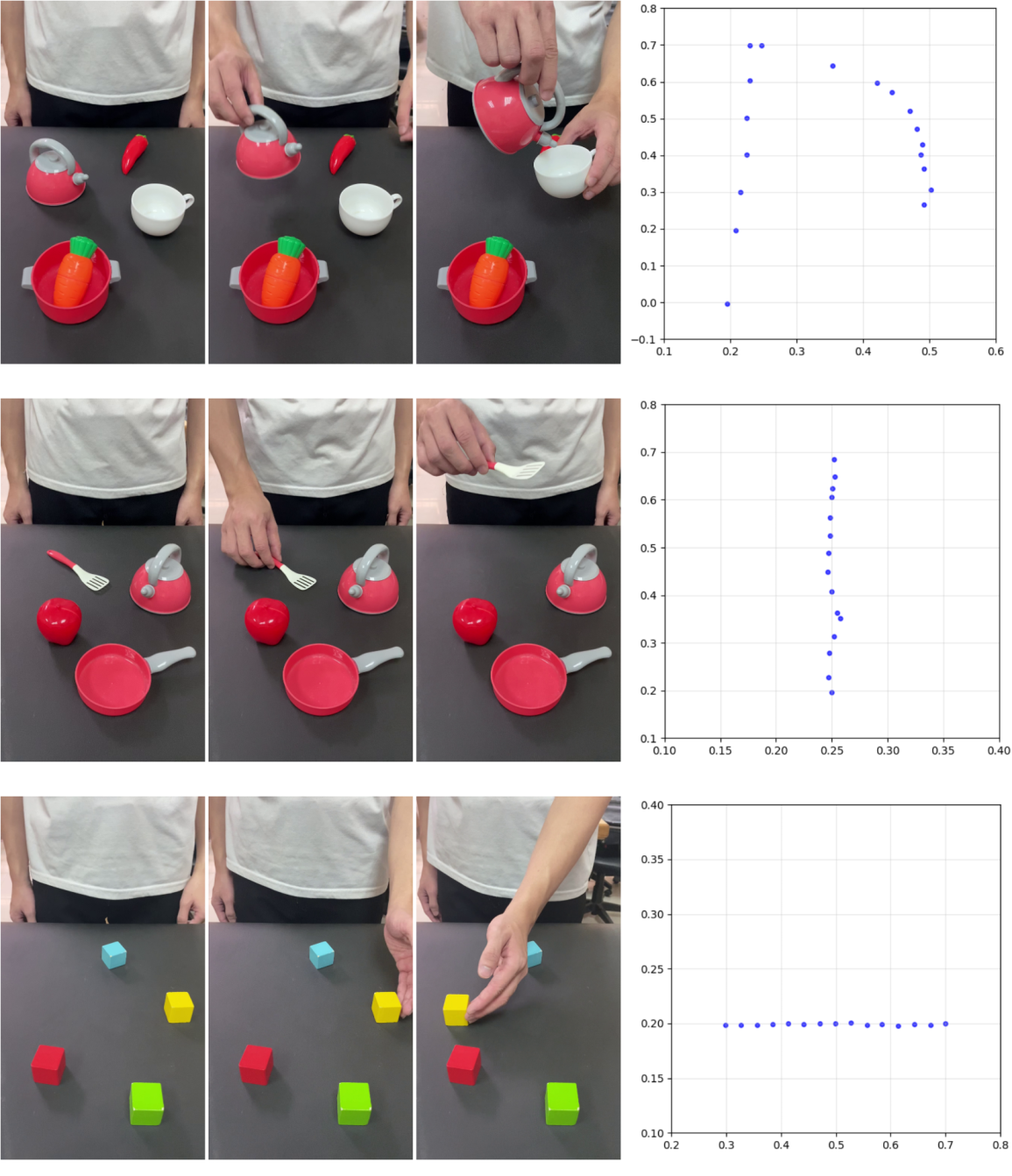}
    \caption{Visualization of Interacted object trajectory tracking across three intersection scenarios. Left: RGB frames showing human hand-object interactions with objects. Right: Corresponding 2D trajectory plots of the tracked objects' centroid positions over time, where the characteristic parabolic or oscillatory patterns indicate pickable objects, enabling automatic classification of object functional roles (pickable or placeable).}
    \label{fig:objtracktrajectory}
\end{figure*}

The standardized, cropped object features ($\mathbf{I}$) are then analyzed to classify their functional role based on the manipulation's context. We define two temporary storage sets to categorize objects by functional role: Pickable Objects ($\mathbf{P}_1$) and Placeable Objects ($\mathbf{P}_2$), as shown in Fig.~\ref{fig:object_selection}. This categorization relies on analyzing the most critical state of the demonstration video, often identifying objects by their motion trajectories, as shown in Fig.~\ref{fig:objtracktrajectory}. Based on these characteristics, we classify objects with a \textit{pickable} functionality into the $\mathbf{P}_1$ group, while objects with a \textit{placable} functionality are assigned to the $\mathbf{P}_2$ group. For refinement, the algorithm applies two primary filters to the object candidates: Blurriness Detection and Overlap Minimization.  Blurriness detection leverages the Laplacian filter matrix  applied $\Bigg ( \mathbf{L} = \begin{bmatrix}
0 & 1 & 0 \\
1 & -4 & 1 \\
0 & 1 & 0
\end{bmatrix} \Bigg )$ via the Window Sliding~\cite{bansal2016blur} principle to compute the blurriness of objects:
\begin{equation}
\mathbf{B}(i, j) = \sum_{k=-1}^{1} \sum_{l=-1}^{1} {\mathbf{i}}_t(i+k, j+l) \cdot \mathbf{L}(k+1, l+1),
\end{equation}
where $\mathbf{i}_{t}(i, j)$ represents the pixel intensity of a cropped image within the set $\mathbf{I}$, and the window size is $3 \times 3$. Objects with the lowest blurriness are retained, as higher blurriness indicates a loss of key features. The second sub-function computes the degree of overlap between the human hand and the interacting object. The objective is to ensure that the selected remains clearly visible in the video without being occluded by other elements. By selecting objects that exhibit both minimal blur and minimal overlap, the system ensures clear, high-quality visual features necessary for accurate grounding of manipulation tasks. These top-ranked object features are then input into VLMs~\cite{liu2023visual} to enhance object understanding and improve generalization across diverse scenarios by integrating textual and visual information. 


\subsubsection{Action Understanding Module (AUM)}
The AUM focuses on identifying the fundamental manipulation primitive performed by the human demonstrator, which corresponds to the ``action'' component ($\mathcal{A}_i$ of the executable command $\mathcal{S}_i$. This module is explicitly designed to learn the spatial-temporal dynamics necessary to classify ten fundamental manipulation actions. In video captioning tasks generally, visual features are extracted offline from video frames using a pre-trained CNN. To prepare the data efficiently for the AUM, the original high frame-rate video sequence $\mathbf{F}$ is first processed into a smaller subset $\tilde{\mathbf{F}}$. This reduction minimizes computational impact by removing duplicate or minimally altered frames, which significantly decreases processing time while maintaining sufficient performance for action classification. Furthermore, it minimizes the gap between training and testing video frame rates, ensuring better consistency.

To accurately capture the temporal properties and fine-grained motion dynamics necessary to distinguish between subtle manipulation actions (\textit{e.g.}, distinguishing ``opening'' from ``closing''), the AUM integrates a Temporal Shift Module (TSM) within its underlying CNN architecture. TSM is a novel and effective mechanism that achieves strong spatio-temporal modeling by sequentially applying random shifts to the channels of the feature sequences. This shift operation facilitates information exchange among neighboring frames along the temporal dimension, allowing the model to fuse temporal information without incurring additional computation or parameters, effectively operating as a pseudo-3D model while maintaining the efficiency of 2D CNNs. Crucially, TSM is implemented using a residual shift strategy, inserting the module inside the residual branch of the network. This structural choice is necessary to prevent performance degradation and preserve the spatial feature learning capacity of the original 2D CNN backbone, which is otherwise harmed by naive "in-place" shift implementations. By leveraging TSM, the AUM efficiently models the complex temporal dependencies crucial for fine-grained human-object interaction tasks, leading to improved performance compared to purely 2D CNN baselines.

\subsubsection{Captioning Scheme}
The final step in the Video Understanding pipeline is the Captioning Scheme, which generates the precise, executable command sentence ($\mathcal{S}_i$) by leveraging the outputs from the AUM and IOUM. This process is vital for providing explicit, structured instructions that directly inform the subsequent Robot Understanding stage. The generated command is produced using a fusion architecture that combines the highest-probability action vectors ($A_k$) from the action understanding phase and the object category vectors ($I_j$) from the interacted-objects understanding phase. By applying a softmax layer to the fused inputs from $\mathcal{A}_k$ and $\mathcal{I}_j$, the final predicted sentence ($\mathcal{S}_i$) is obtained. The structured output command is defined by the concatenation of these two semantic components, emphasizing the action and the target objects: 
\begin{equation}
    \mathcal{S}_i = \mathcal{A}_i + \mathcal{I}_i, \quad i = 1, \dots, m
\end{equation}

The resulting command sentence $S_i$ is explicitly enforced to be in a grammar-free format (\textit{e.g.}, ``place orange in bowl'' instead of complex natural language descriptions) for the convenience of robot execution. This concise format emphasizes the essential components—the action and the target object—ensuring clarity and direct applicability for downstream robotic manipulation tasks. In experiments, the number of output words can reach up to 8, and is typically set to a maximum length by padding the remaining list with empty words, as generated command sentences generally consist of less than this limit

\subsection{Robot Imitation}

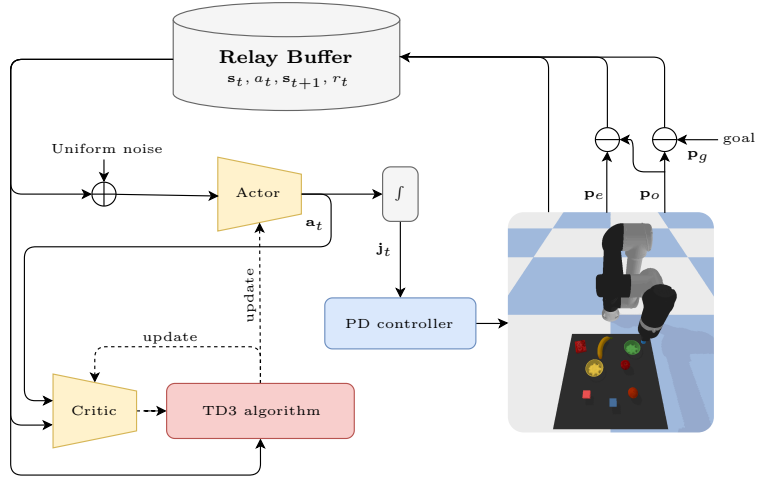
\begin{figure*}[!ht]
\centering
    \begin{overpic}[width=0.8\textwidth, unit=1pt]{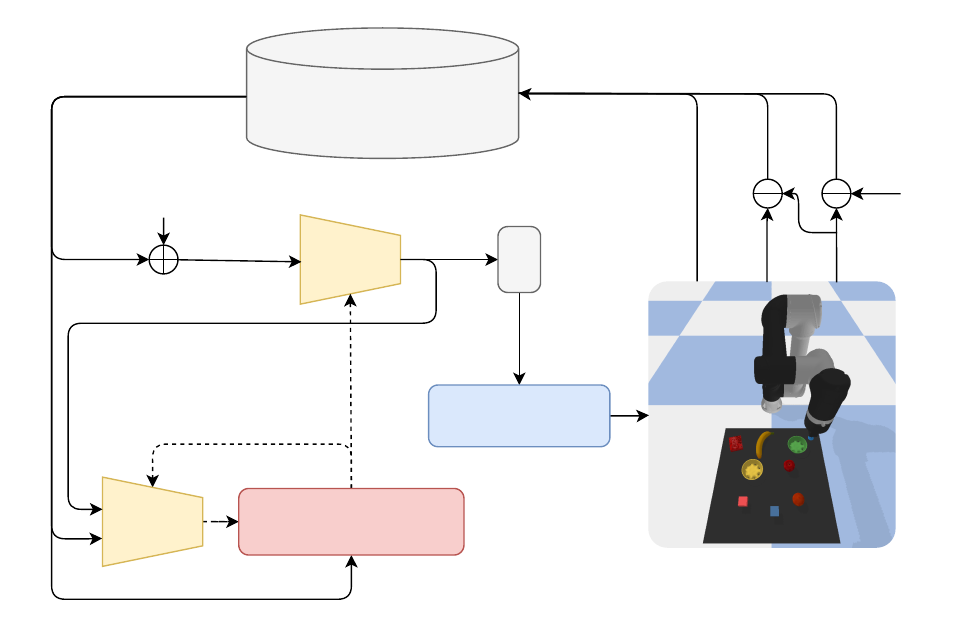}
        \put(13.0, 11.8){\tiny Critic}
        \put(29.5, 11.8){\tiny TD3 algorithm}
        \put(47.5, 22.8){\tiny PD controller}
        \put(22.0, 21.2){\tiny update }
        \put(34.8, 23.5){\tiny \rotatebox{90}{update}}
        \put(10.5, 45.0){\tiny Uniform noise}
        \put(33.7, 39.5){\tiny Actor}
        \put(42.5, 35.8){\tiny $\mathbf{a}_t$}
        \put(51.5, 32.5){\tiny $\mathbf{j}_t$}
        \put(53.7, 39.5){\tiny $\int$}
        \put(31.6, 56.5){\footnotesize \textbf{Relay Buffer}}
        \put(32.7, 54.0){\tiny $\mathbf{s}_t, a_t, \mathbf{s}_{t+1}, r_t$}
        \put(77.5, 39.5){\tiny $\mathbf{p}_e$}
        \put(84.5, 39.5){\tiny $\mathbf{p}_o$}
        \put(90.5, 44.5){\tiny $\mathbf{p}_g$}
        \put(95.0, 46.5){\tiny goal}
    \end{overpic}
\caption{The architecture of our DRL system for robot imitation. processes the environment observation $\mathcal{O}$ that consists of agent's proprioceptive, exteroceptive, relational, and historical data such as: joint angle $\mathbf{j}_{t}$, joint velocity $\mathbf{j}^\prime_{t}$, object orientation $\mathbf{R}_{o}$, object coordination and target $\mathbf{p}_{o}$, $\mathbf{p}_{g}$, end effector status $\mathbf{p}_{e}$, we also provide relative observations $\bar{\mathbf{p}}_{eo}$, $\bar{\mathbf{p}}_{og}$ with past joint angle \(\mathbf{a}_{t-1}\), provided by the Video Understanding framework and outputs the location of the object referenced by \(\mathbf{i}_t\).}
\label{fig:drl_pipeline}
\end{figure*}


\subsubsection{State and Action Representation}

Effective decision-making requires a comprehensive state embedding that captures both robot kinematics and task-relevant environmental cues. To enhance sample efficiency and generalization, we adopt domain randomization on both state representation (Table~\ref{table:state_representation}) and environment configuration, maintaining 100\% randomness while ensuring reproducibility. Table~\ref{table:state_representation} also summarizes each component of the state vector $\mathbf{s}_t$, which includes:
\begin{itemize}
    \item Proprioceptive information: Joint angles $\mathbf{j}_t$ and joint velocities $\mathbf{j}^\prime_{t}$ provide essential feedback for accurate motion control and trajectory planning.
    \item Spatial Relationships: Relative position vectors $\bar{\mathbf{p}}_{eo}$ (end-effector to object) and $\bar{\mathbf{p}}_{og}$ (object to goal) encode the spatial relationships critical for approach and manipulation phases. Object position $\mathbf{p}_o$, goal position $\mathbf{p}_g$, and object orientation $\mathbf{R}_o$ complete the spatial description.
    \item Interaction State: Suction values $\mathbf{p}_e$ comprise two binary indicators: one reflecting the activation/release state and another representing the predicted value from the policy, enabling the agent to reason about grasp success.
    \item Temporal Context: Including the previous actions $\mathbf{a}_{t-1}$ helps the agent learn smooth motion trajectories by mitigating discontinuities in control signals.
\end{itemize}

\begin{table}[!ht]
\caption{State representation for the DRL agent.}
\centering
\begin{tabularx}{\textwidth} { 
  | >{\centering\arraybackslash}X 
  | >{\centering\arraybackslash} p{0.3\textwidth} 
  | >{\centering\arraybackslash}X
  | >{\centering\arraybackslash}X |}
\toprule
\textbf{Component}     & \textbf{Description}         & \textbf{Dimension} & \textbf{Noise} \\ \midrule
$\mathbf{j}_{t}$       & joint positions at time $t$  & 6D & $\mathcal{U}(\pm 0.005)$ \\ 
$\mathbf{j}^\prime_{t}$ & joint velocities at time $t$ & 6D & $\mathcal{U}(\pm 0.05)$  \\
$\mathbf{R}_{o}$       & object orientation           & 4D & $\mathcal{U}(\pm 0.005)$ \\ 
$\bar{\mathbf{p}}_{eo}$      & end-effector to object vector  & 3D & $\mathcal{U}(\pm 0.005)$ \\
$\bar{\mathbf{p}}_{og}$      & object to goal vector        & 3D & $\mathcal{U}(\pm 0.005)$ \\
$\mathbf{p}_{o}$       & object coordination          & 3D & $\mathcal{U}(\pm 0.005)$ \\
$\mathbf{p}_{g}$       & target coordination          & 3D & $\mathcal{U}(\pm 0.005)$ \\
$\mathbf{p}^e_{t}$     & suction sensor readings      & 2D & $\mathcal{U}(\pm 0.005)$ \\
$\mathbf{a}_{t-1}$     & previous joint command       & 7D & ---                        \\

\bottomrule
\end{tabularx}%
\label{table:state_representation}
\end{table}

To ensure stable training, we normalize observations $\mathbf{O}_t \in \mathcal{O}$ using Welford's online algorithm, which maintains running estimates of mean $\hat{\mu}$ and variance $\sigma^2$:
\begin{equation}
    \begin{aligned}
    M_{2,t} = M_{2,{t-1}} + (\mathbf{O}_t - \hat{\mu}_{t-1}) \odot (\mathbf{O}_t - \hat{\mu}_t), \qquad \sigma_t^2 = \frac{M_{2,t}}{n_t}
    \end{aligned}
\end{equation}
where $n_t = n_{t-1} + 1$ is the number of samples seen, and $M^2_t$ accumulates the sum of squared deviations. The normalized observation (standardization) is computed as:
\begin{equation}
\hat{\mathbf{O}} = 
\frac{\mathbf{O} - \hat{\mu}_t}{\sigma_t + \epsilon}, \qquad \epsilon = 1e-7
\end{equation}

The action $\mathbf{a}_t$ specifies two components: continuous joint velocities for all robot joints and a discretized suction signal modeled via a Bernoulli distribution. Joint positions are obtained by integrating predicted velocities:
\begin{equation}
\mathbf{j}_t = \int \mathbf{j}^\prime_{t} \,dt= \mathbf{j}_{t-1} + \mathbf{j}^\prime_{t} \cdot dt
\label{eq:controller_equation}
\end{equation}

These commanded positions are then regulated by a low-level PD controller that enforces velocity and joint-limit constraints. This control scheme extends beyond the standard Markov Decision Process (MDP) assumption, as the state transition depends on both current and previous actions: $P(\mathbf{s}_{t+1}/ \mathbf{a}_t, \mathbf{s}_t, \mathbf{a}_{t-1})$.

\subsubsection{Reward \& Penalty Function}
\begin{table*}[!ht]
\centering
\footnotesize
\caption{Rewards and penalties formula used for training, along with the initial and final weights.  \\ \footnotesize{$^*$Notation: $\mathbf{p}_e$ is end-effector position; $\lambda_\_$ is scale factor; $\theta_e, \theta_o, \psi_e, \psi_o$ are roll and pitch of end-effector and object respectively; $p_{\_}^x, p_{\_}^y, p_{\_}^z$} are position in x, y, z-axis respectively; $\theta_o^{c}$ is a constraint; $\mathbf{w}^{c}, w_x^c, w_y^c$ are position constraint of workspace in xyz coordinate, only x-axis, and only y-axis respectively.}
\begin{tabularx}{1.0\textwidth} { 
  | >{\raggedright\arraybackslash} m{0.55\textwidth} 
  | >{\arraybackslash}m{0.18\textwidth}
  | >{\centering\arraybackslash} m {0.17\textwidth}| }
\toprule
\centering \textbf{Rewards/Penalties} & \centering \textbf{Description} & \textbf{Weight} \\
\midrule
$\begin{aligned}
    r^e_{t} = w_0\exp\!\left(-\frac{\lVert{}\mathbf{p}_e - \mathbf{p}_o\rVert}{\lambda_0}\right) &+ w_1\exp\!\left(-\frac{|\psi_e-\pi| + |\theta_e|}{\lambda_1}\right) \\ &+  w_2\exp\!\left(-\frac{\lVert{}p^z_{e} - p^z_{o}\rVert}{\lambda_2}\right)
\end{aligned}$ 
& safe end-effector approach to the object & $(1.3,0.2, 0.15) \rightarrow (0.325, 0.2, 0.15)$ \\ \hline

$\begin{aligned}
    r^i_{t} =  &\delta_c \left(w_3\exp\!\left(-\frac{\lVert{}\mathbf{p}_o - \mathbf{p}_g\rVert}{\lambda_3}\right) + w_4\exp\!\left(-\frac{\lVert{}\mathbf{p}_o - \mathbf{p}_g\rVert}{\lambda_4}\right)\right), \\ &\delta_c = 1 \text{ if contact is established}, \qquad \lambda_3 >> \lambda_4
\end{aligned}
$  
& object movement toward the target & $(2.0, 3.0)$ \\ \hline

$r^a_{t} = w_5\exp\!\left(\frac{\bar{\mathbf{p}}^\prime_o \cdot\bar{\mathbf{p}}_{og}}{\lambda_5} - 1\right) \bigg[\bar{\mathbf{p}}^\prime_o \cdot\bar{\mathbf{p}}_{og} > 0\bigg ]$ 
& object motion aligned with goal direction & 0.8 \\ \hline
$c_{c} =
\begin{cases}
3, & \text{if colliding with ground or itself} \\ 1.5, & \text{if colliding with objects} \\ 0, & \text{otherwise.}
\end{cases}$ & undesired collisions   & 1.0 \\ \hline
$c_{s} = 2$ & reaching the max step limit & 1.0 \\ \hline
$c_{e} = w_6\left(|\psi_e-\pi| - |\theta_e|\right)$ & end-effector pose inclination limitation & 1.0 \\ \hline
$c_{a} = w_7\left(\lVert{}\mathbf{a}_t - 2\mathbf{a}_{t-1} + \mathbf{a}_{t-2}\rVert\right)$ & jerky or unsmooth actions limitation & $0.05$ \\ \hline
$c_{o} =
\begin{cases}
\left(|\psi_o - \psi_o^{c}| - |\theta_o - \theta_o^{c}|\right), & \text{if } \psi_o \text{ or } \theta_o > \epsilon_t, \\
0, & \text{otherwise.}
\end{cases}$ & excessive object inclination limitation & 1.0 \\ \hline

$c_{w} =
\begin{cases}
\frac{\lVert \mathbf{p}_o - \mathbf{w}^c \rVert + \lVert \mathbf{p}_e - \mathbf{w}^c \rVert}{\sigma_1}
, \text{if } &((p^x_{o} \cup  p^x_{e}) \notin  w_x^c)) \cup \\ 
             &((p^y_{o} \cup  p^y_{e}) \notin  w_y^c) \\
0, & \text{otherwise.}
\end{cases}$ & movement outside workspace limitation & 1.0 \\ \bottomrule
\end{tabularx}
\label{table:reward_penalty}
\end{table*}
The primary objective of reinforcement learning is to maximize the cumulative discounted reward, which guides the agent toward optimal behavior. Designing an effective reward function is critical for aligning the agent's actions with task requirements. In this work, the reward function is formulated as a weighted combination of multiple components, each targeting specific aspects of manipulation to encourage desired behaviors while penalizing errors.
The total reward at timestep $t$ is computed as:
\begin{equation}
r_{t}(\mathbf{s}_t, \mathbf{a}_t) = \sum_{i=0}^{2} w_i \cdot r^i_{t}(\mathbf{s}_t, \mathbf{a}_t) - \sum_{j=1}^{6} w_j \cdot c^j_{t}(\mathbf{s}_t, \mathbf{a}_t),
\label{eq:total_reward}
\end{equation}
where $r^i_t$ denotes reward terms promoting task progress, $c^j_t$ denotes penalty terms discouraging unsafe behaviors, and $w_i$, $w_j$ are their respective weights. To ensure gradient stability during critic training, we normalize the total reward $r_t$ by a scale factor of 10.

\paragraph{Reward Components:}
We employ a hierarchical reward structure that provides clear feedback based on observable action outcomes, guiding the agent through sequential sub-goals to reach or interact with the objects. The reward terms are defined as follows:

\begin{itemize}
    \item Approach Reward ($r^e_{t}$): Encourages the end-effector to safely approach the target object with proper positioning and orientation. This term uses exponential distance-based shaping to provide smooth gradients throughout the approach phase.
    
    \item Interaction Reward ($r^i_{t}$): Activated upon contact with the target object, this term incentivizes transporting the grasped object toward the goal location. The reward is conditioned on successful grasp detection via the \texttt{has\_contact} indicator.
    
    \item Alignment Reward ($r^a_{t}$): Rewards object motion that is aligned with the goal direction, computed as the dot product between the object velocity vector $\bar{\mathbf{p}}^\prime_o$ and the object-to-goal unit vector $\bar{\mathbf{p}}_{og}$. This term is only activated when the object is moved ($\bar{\mathbf{p}}^\prime_o \cdot \bar{\mathbf{p}}_{og} > 0$).
\end{itemize}

Upon successful task completion, we set $r^i_{t} = r^i_{t-1}$ (indicating no further movement is required), $r^e_{t} = 0$, and award a triple maximum score for $r^a_{t}$ to reinforce successful completion. This hierarchical structure ensures that the policy learns to execute each manipulation phase precisely before progressing to the next.

\paragraph{Penalty Components:}
In contrast to rewards that promote goal-directed behavior, penalties ensure safe execution by discouraging undesirable actions such as collisions, joint limit violations, and workspace boundary crossings. The penalty terms are defined as follows:

\begin{itemize}
    \item Collision Penalty ($c_{c}$): Penalizes undesired collisions with graduated severity—ground or self-collisions receive the highest penalty (3.0), while collisions with scene objects receive a moderate penalty (1.5).
    
    \item Step Limit Penalty ($c_{s}$): Applied when the episode reaches the maximum allowed timesteps, encouraging efficient task completion.
    
    \item End-Effector Inclination Penalty ($c_{e}$): Discourages excessive tilting of the end-effector, maintaining proper orientation for stable manipulation.
    
    \item Action Smoothness Penalty ($c_{a}$): Penalizes jerky or discontinuous motions by minimizing the second derivative of the action sequence ($\|a_t - 2a_{t-1} + a_{t-2}\|$), promoting smooth trajectories.
    
    \item Object Inclination Penalty ($c_{o}$): Activated when the manipulated object tilts beyond a predefined threshold $\epsilon_t$, preventing unstable grasps that could lead to drops.  
    
    \item Workspace Penalty ($c_{w}$): Penalizes movements that would place the end-effector or object outside the defined workspace boundaries, ensuring safe operation within the robot's operational envelope.
\end{itemize}

The complete mathematical formulations for all reward and penalty terms, along with their associated weights, are provided in Table~\ref{table:reward_penalty}. The inclusion of adaptive penalties, particularly for critical failures, helps balance exploration with task-focused exploitation during training.

\subsubsection{Training Procedure}

We utilize a deep reinforcement learning (DRL) framework based on the TD3 algorithm~\cite{fujimoto2018addressing}, illustrated in Fig.~\ref{fig:drl_pipeline}. We employ deterministic policies rather than stochastic alternatives because, in highly randomized environments, stochastic policies tend to inefficiently adjust their mean and variance parameters, leading to increased gradient variance and unstable learning dynamics. The critic networks are trained by minimizing the temporal difference (TD) error between the estimated Q-values and the target values. For the $i$-th critic network ($i \in \{1, 2\}$), the loss function is defined as:
\begin{equation}
\mathcal{L}_{\phi_i} = \mathbb{E}_{(\mathbf{s}_t, \mathbf{a}_t, r_t, \mathbf{s}_{t+1}) \sim \mathcal{B}} \left[ \left( Q_{\phi_i}(\mathbf{s}_t, \mathbf{a}_t) - y_t \right)^2 \right], \quad i = 1, 2
\label{eq:critic_loss}
\end{equation}
where $\mathcal{B}$ denotes the replay buffer, and the target value $y_t$ is computed using the minimum of the two target critic networks to prevent overestimation:
\begin{equation}
y_t = r_t + \gamma \cdot \min_{i=1,2} Q_{\phi'_i}\left(\mathbf{s}_{t+1}, \pi_{\theta'}(\mathbf{s}_{t+1}) + \epsilon\right), \quad \epsilon \sim \text{clip}\left(\mathcal{N}(0, \tilde{\sigma}), -c, c\right)
\label{eq:target_value}
\end{equation}
where, $\gamma$ is the discount factor, $\phi'_i$ and $\theta'$ denote the target network parameters, and $\epsilon$ is clipped noise added for target policy smoothing.

The actor network is updated by maximizing the expected Q-value through gradient ascent. Following the deterministic policy gradient theorem, the actor loss is:
\begin{equation}
\mathcal{L}_{\theta} = -\mathbb{E}_{\mathbf{s}_t \sim \mathcal{B}} \left[ Q_{\phi_1}(\mathbf{s}_t, \pi_{\theta}(\mathbf{s}_t)) \right]
\label{eq:actor_loss}
\end{equation}
The actor is updated at a lower frequency than the critics (every $d$ iterations) to ensure stable Q-value estimates before policy improvement. During training, each episode begins by randomly generating a scene containing 19 different objects sampled from the training object set. For the \textit{reach} action, the agent must navigate its end-effector to the target object position. For \textit{pick}, \textit{move}, and \textit{put} actions, the agent must accomplish two sequential sub-goals: (1) reaching and grasping the correct target object, and (2) transporting it to the designated goal location. In addition, an episode terminates when any of the following conditions is satisfied:
\begin{itemize}
    \item Success: The agent successfully completes the required action with the target object within the specified tolerance ($\|p_o - p_g\| \leq \tau_d$).
    \item Collision: The end-effector collides with the ground plane or the robot's own body.
    \item Joint limit violation: Any joint angle exceeds its permissible range.
    \item Workspace violation: The object or end-effector moves outside the defined workspace boundaries.
    \item Timeout: The episode exceeds the maximum allowed timesteps ($T_{max} = 100$).
\end{itemize}

The complete training pipeline, integrating video understanding with reinforcement learning, is summarized in Algorithm~\ref{alg:video_imitation_rl}.

\begin{algorithm}[!ht]
\caption{Human-to-Robot Imitation Learning Pipeline}
\label{alg:video_imitation_rl}
\begin{algorithmic}[1]
\small
\Require Video demonstrations $\mathcal{D}$, robot environment $\mathcal{E}$
\Ensure Learned policy $\pi_\theta$

\State Initialize modules: $\mathcal{I}$, $\mathcal{A}$, $\pi_\theta$, $Q_{\phi_1}$, $Q_{\phi_2}$
\State Initialize targets: $\theta' \leftarrow \theta$, $\phi'_i \leftarrow \phi_i$
\State Initialize replay buffer: $\mathcal{B} \leftarrow \emptyset$

\Comment{\textbf{Phase 1: Video Understanding}}
\For{each $\mathcal{V} \in \mathcal{D}$}
    \State $\tilde{\mathbf{F}} \leftarrow \text{Downsample}(\mathcal{V})$ \Comment{Reduce frame rate}
    \State $A \leftarrow \mathcal{A}(\tilde{F})$ \Comment{Action classification}
    \State $\hat{\mathbf{F}} \leftarrow \text{KeyframeExtract}(\tilde{\mathbf{F}})$ \Comment{Select keyframes}
    \State $O \leftarrow \mathcal{O}(\hat{F})$ \Comment{Object identification}
    \State $S \leftarrow \text{Concat}(A, O)$ \Comment{Generate command}
\EndFor

\Comment{\textbf{Phase 2: Reinforcement Learning}}
\For{episode $= 1$ to $M$}
    \State $s_0 \leftarrow \mathcal{E}.\text{reset}(S)$
    \For{$t = 0$ to $T_{max}$}
        \State $a_t \leftarrow \pi_\theta(s_t) + \epsilon$, \quad $\epsilon \sim \mathcal{N}(0, \sigma)$ \Comment{Exploration}
        \State $(s_{t+1}, r_t, done) \leftarrow \mathcal{E}.\text{step}(a_t)$
        \State $\mathcal{B} \leftarrow \mathcal{B} \cup \{(s_t, a_t, r_t, s_{t+1})\}$
        \State Update $Q_{\phi_1}, Q_{\phi_2}$ via Eq.~\eqref{eq:critic_loss} \Comment{Critic update}
        \If{$t \mod d = 0$}
            \State Update $\pi_\theta$ via Eq.~\eqref{eq:actor_loss} \Comment{Delayed actor update}
            \State $\theta' \leftarrow \rho\theta + (1-\rho)\theta'$ \Comment{Soft update}
            \State $\phi'_i \leftarrow \rho\phi_i + (1-\rho)\phi'_i$
        \EndIf
        \If{$done$}
            \State break
        \EndIf
    \EndFor
\EndFor

\State \textbf{Return:} $\pi_\theta$

\end{algorithmic}
\end{algorithm}

\begin{figure}[!ht]
\centering
\begin{subfigure}[!ht]{0.85\textwidth}
    \centering
    \includegraphics[width=\textwidth]{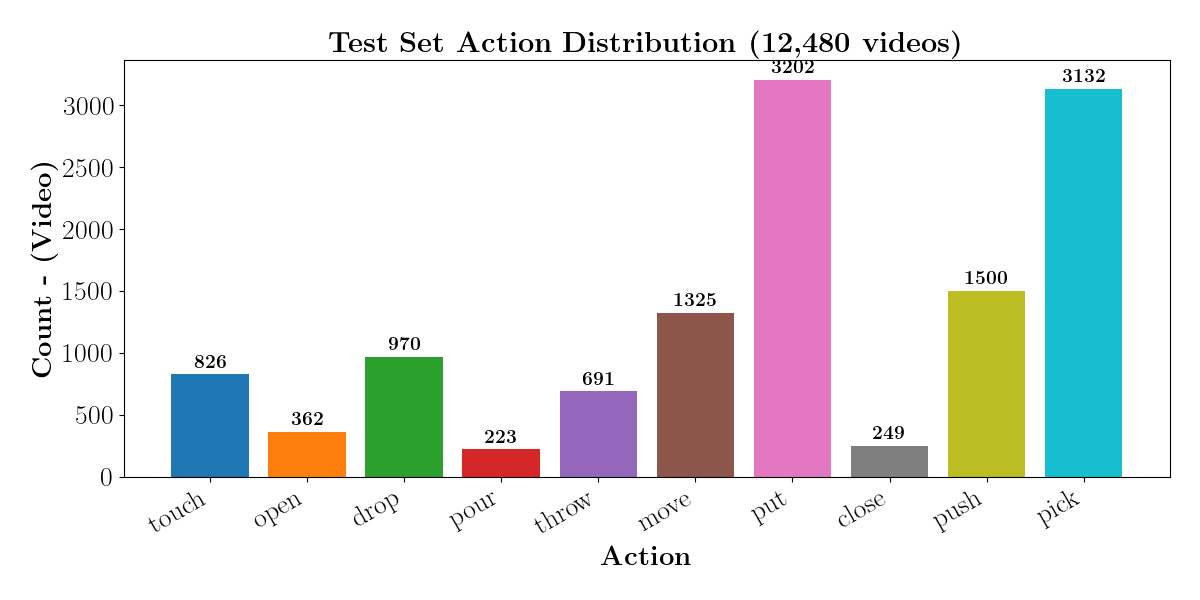}
    \caption{Original dataset}
\end{subfigure}
\hfill
\begin{subfigure}[!ht]{0.85\textwidth}
    \centering
    \includegraphics[width=\textwidth]{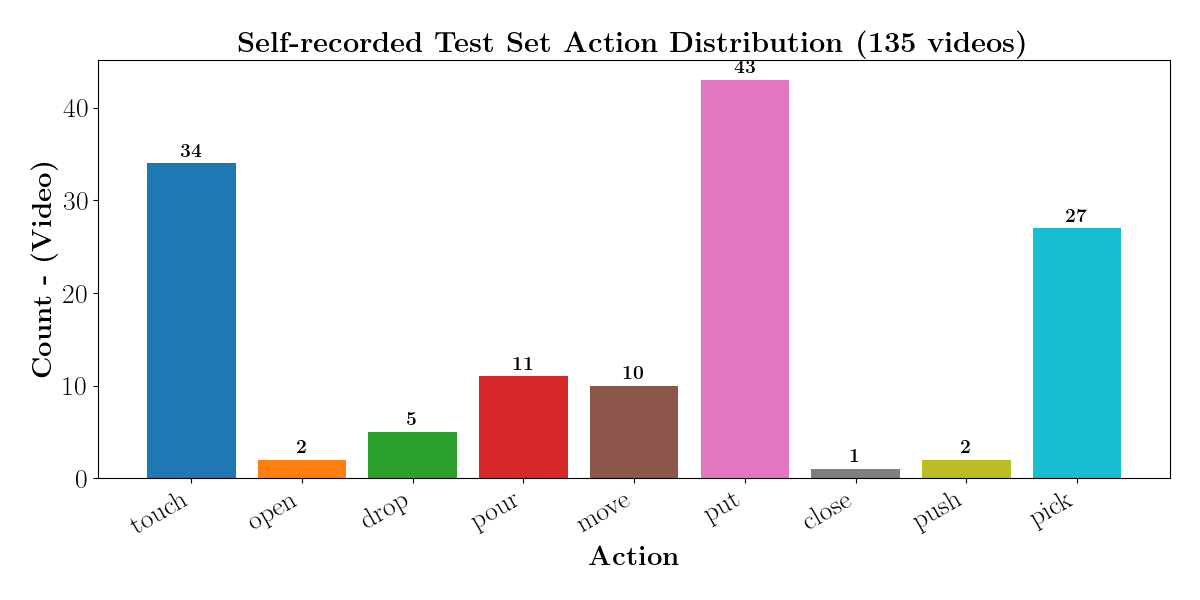}
    \caption{Our dataset}
\end{subfigure}
\caption{Distribution of training videos across action categories in the test set}
\label{fig:dataset_distribution}
\end{figure}

\begin{figure}[!ht]
\centering
    \includegraphics[width=\textwidth]{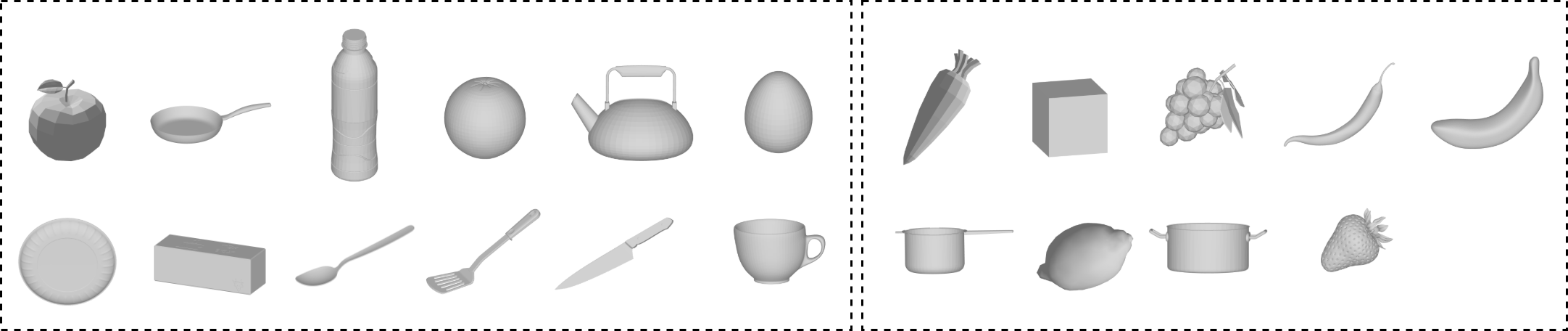}
\caption{Object categories used for evaluating video understanding generalization. (Left) Standard object set: 12 object categories present in the training dataset, including common household items. (Right) Novel object set: 9 previously unseen object categories used exclusively for testing}
\label{fig:objects}
\end{figure}

\begin{figure*}[!ht]
    \centering
    \includegraphics[width=\textwidth]{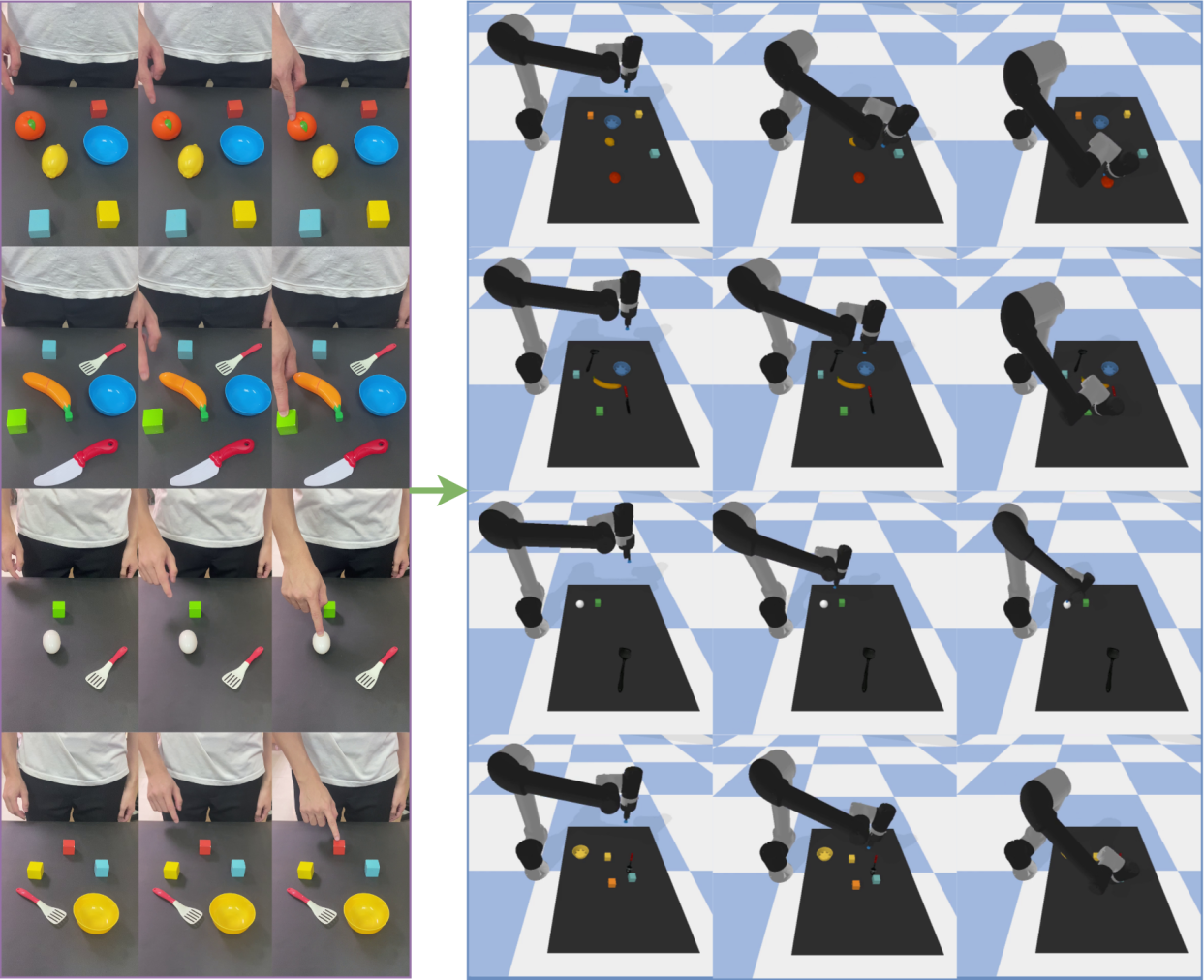}
    \caption{Qualitative results of reach action. Left: Human demonstration videos showing reaching motion toward various objects. Right: Corresponding simulated UR5 robot executions, where the robot successfully navigates its end-effector to the target object position across diverse scene configurations with multiple distractor objects.}
    \label{fig:qual_reach}
\end{figure*}

\begin{figure*}[!ht]
    \centering
    \includegraphics[width=\textwidth]{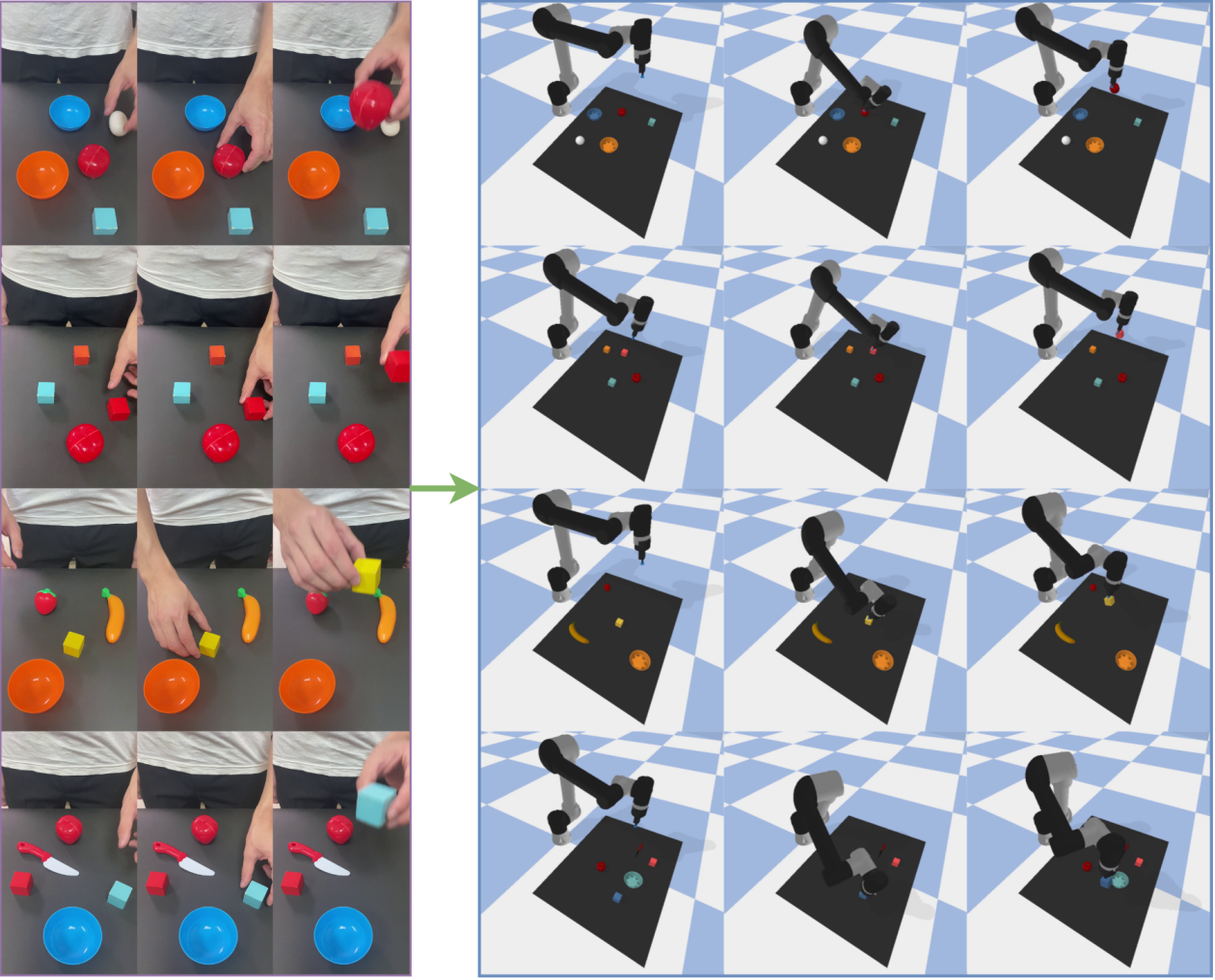}
    \caption{Qualitative results of the pick action. Left: Human demonstration videos showing grasping motions for objects of varying colors and shapes. Right: Corresponding simulated executions using the suction gripper, with the robot correctly identifying and picking the target object among multiple distractors.}
    \label{fig:qual_pick}
\end{figure*}

\begin{figure*}[!ht]
    \centering
    \includegraphics[width=\textwidth]{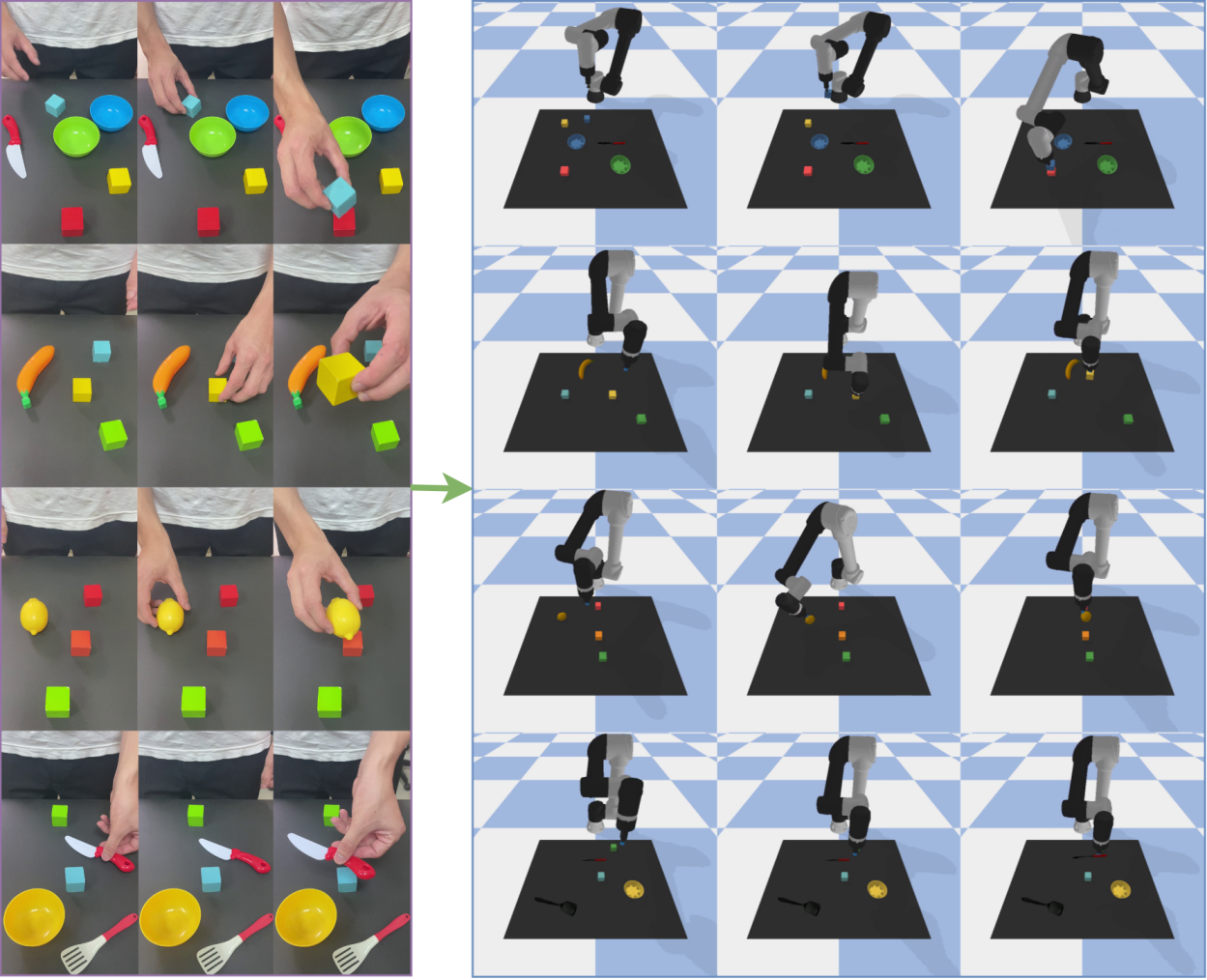}
    \caption{Qualitative results of the move action. Left: Human demonstration videos showing object transportation from one location to another. Right: Corresponding simulated executions, where the robot successfully transports grasped objects along trajectories toward the designated goal positions while maintaining a stable grasp throughout the motion.}
    \label{fig:qual_move}
\end{figure*}

\begin{figure*}[!ht]
    \centering
    \includegraphics[width=\textwidth]{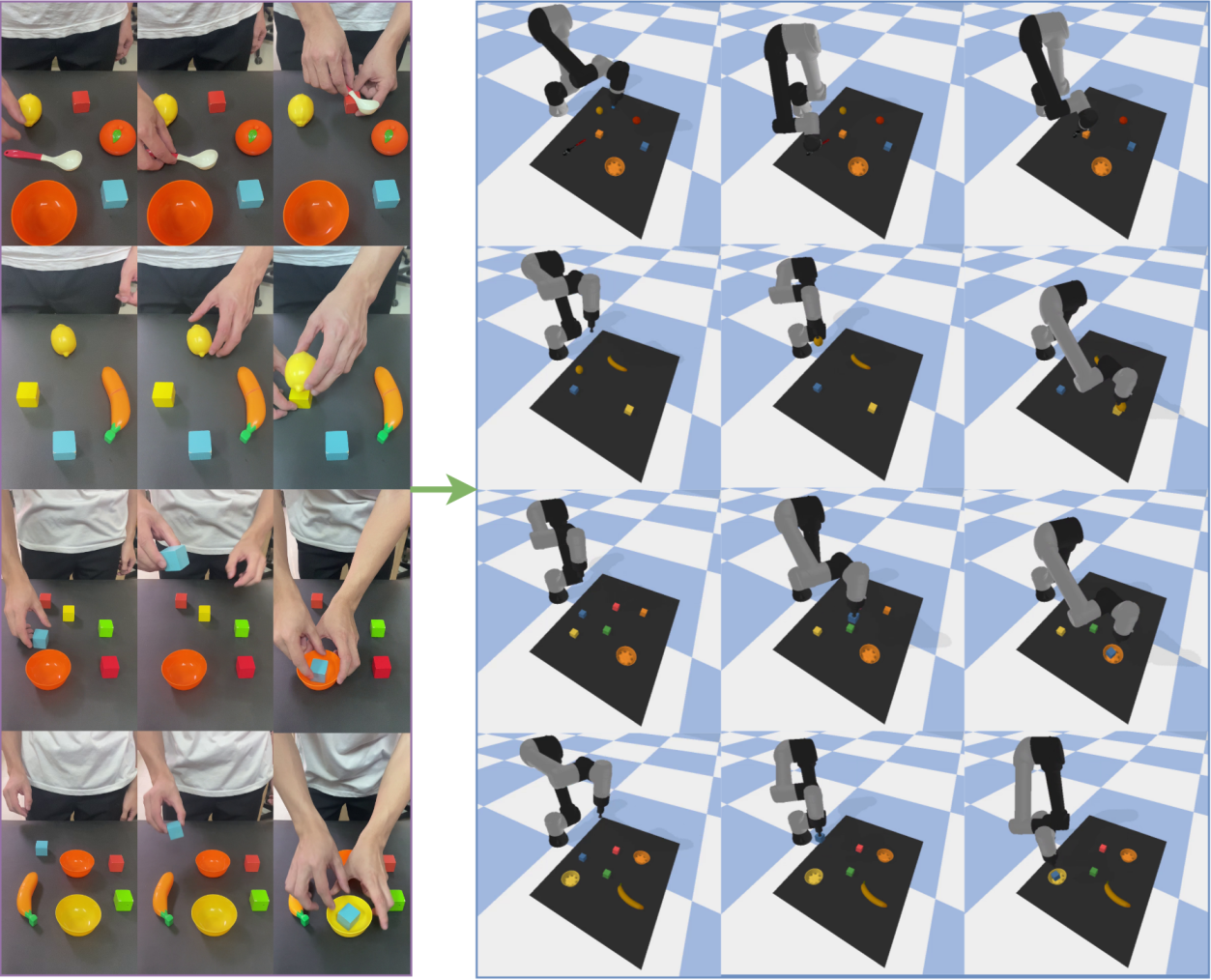}
    \caption{Qualitative results of the put action. Left: Human demonstration videos showing object placement into containers (bowls, plates). Right: Corresponding simulated execution demonstrating precise placement of objects at target locations, completing the full pick-and-place manipulation sequence learned from video demonstration.}
    \label{fig:qual_put}
\end{figure*}

\section{Results and Discussion}
\label{sec:result}
\subsection{Experimental Setup}
We evaluate the proposed framework through comprehensive simulation experiments, measuring task success rates across diverse manipulation scenarios with varying object shapes, colors, and spatial configurations. All experiments are conducted in PyBullet~\cite{coumans2016pybullet} using a Universal Robot UR5e manipulator equipped with a suction gripper. This configuration provides a controlled and reproducible platform for evaluating manipulation performance, particularly for tasks requiring semantic grounding of high-level cues such as object color and category. The robot operates within a fixed $60 \times 60$~cm tabletop workspace, with RGB images captured by an onboard camera at $640 \times 480$ resolution without sensor noise. The policy networks are trained on NVIDIA Tesla V100-SXM2-32GB and GeForce RTX 5080-16GB GPUs. Training consists of $35,000$ episodes across randomized environments, totaling $\sim3,500,000$ environment steps. Each iteration requires approximately 2 seconds in simulation, resulting in a total training time of approximately 24 hours. Table~\ref{tab:hyperparameters} summarizes the key hyperparameters used for TD3 training.

\begin{table}[t]
\centering
\caption{DRL training hyperparameters.}
\label{tab:hyperparameters}
\begin{tabularx}{0.6\textwidth} { 
   >{\raggedright\arraybackslash} m{0.25\textwidth}|| 
   >{\centering\arraybackslash} X }
\toprule
\textbf{Hyperparameter} & \textbf{Value} \\
\midrule
Learning rate (actor) & $4 \times 10^{-4}$ \\
Learning rate (critic) & $4 \times 10^{-4}$ \\
Discount factor $\gamma$ & 0.99 \\
Soft update rate $\rho$ & 0.005 \\
Replay buffer size & $10^6$ \\
Batch size & 256 \\
Policy delay $d$ & 2 \\
Exploration noise $\sigma$ & 0.2 \\
Target noise $\tilde{\sigma}$ & 0.2 \\
Noise clip $c$ & 0.5 \\
\bottomrule
\end{tabularx}
\end{table}

\paragraph{Evaluation Metrics:} To assess task performance, we define \textit{success rate} as the percentage of episodes in which the robot completes all required sub-goals within a positional error threshold $\tau_d$. We report success rates under varying thresholds ($\tau_d \in \{1, 2, 3, 4\}$~cm) to evaluate precision-performance trade-offs. Additionally, we measure the \textit{average reward}, defined as the cumulative discounted reward $R = \sum_{t=0}^{T} \gamma^t r_t$ accumulated during an episode, which reflects both task completion and motion quality. For the Video Understanding module, we use \textit{BLEU scores} (BLEU-1 through BLEU-4) to evaluate the accuracy of generated commands against ground-truth annotations.

\paragraph{Baseline Methods:}
We compare our approach against several state-of-the-art methods spanning video-to-command generation and robotic manipulation. For video understanding, we evaluate against Video2Command~\cite{nguyen2018translating}, a deep recurrent neural network approach using sequence-to-sequence learning; V2CNet~\cite{nguyen2019v2cnet}, which incorporates an additional action classification branch; and Watch-and-Act~\cite{yang2023watch}, which utilizes Visual Change Maps with Mask R-CNN for object-centric action recognition. For DRL algorithm comparison, we additionally evaluate against SAC~\cite{haarnoja2018soft}, DDPG~\cite{lillicrap2015continuous}, PPO~\cite{schulman2017proximal}, and Asym-PPO~\cite{jeon2023learning}.

\paragraph{Dataset:}

We train the Temporal Shift Module on a modified Something-Something V2 dataset, comprising videos across 10 action categories. The training set contains $118,562$ videos, while the test set includes $12,480$ videos. Additionally, we collect $135$ self-recorded videos to evaluate real-world generalization (see Fig.~\ref{fig:dataset_distribution}). On the other hand, we evaluate object recognition on two sets: (1) a \textit{standard set} containing 12 object categories present during training (apple, pan, bottle, orange, kettle, egg, plate, box, spoon, spatula, knife, cup), and (2) a \textit{novel set} containing 9 previously unseen categories (carrot, block, grape, chilly, banana, pressure, lemon, pot, strawberry) to assess zero-shot generalization capability (Fig~\ref{fig:objects}).

\subsection{Experimental Results}
We present comprehensive experimental results evaluating both the Video Understanding and Robot Imitation components of our framework. All experiments are conducted over 10 independent trials to ensure statistical reliability.

\subsubsection{Video Understanding Performance}

Fig.~\ref{fig:qual_reach}--\ref{fig:qual_put} present qualitative comparisons between human demonstration videos (left panels) and corresponding simulated robot executions (right panels) across four manipulation primitives. Fig.~\ref{fig:qual_reach} illustrates the \textit{reach} action across six diverse scenarios involving spherical, cylindrical, and irregular objects. The UR5e successfully navigates to target positions despite cluttered scenes with 5--8 distractors, correctly identifying targets even among visually similar objects and maintaining smooth, collision-free trajectories. Fig.~\ref{fig:qual_pick} demonstrates the \textit{pick} action, where the robot must establish stable contact for grasping. The suction gripper successfully identifies and picks targets among distractors, consistently approaching from above with proper orientation as encouraged by $r^e_t$. Failed cases primarily occur near workspace boundaries or when objects are in close proximity. Fig.~\ref{fig:qual_move} showcases the \textit{move} action for object transportation. The robot exhibits smooth trajectories, lifting objects to safe heights before horizontal translation. The alignment reward $r^a_t$ guides efficient near-linear paths, while penalties $c_e$ and $c_o$ maintain end-effector orientation and object stability throughout transportation. Fig.~\ref{fig:qual_put} presents the complete pick-and-place sequence through the \textit{put} action. The robot achieves precise placement into containers, with the interaction reward $r^i_t$ guiding the final approach phase. The consistently high success rates (90\% at 1~cm threshold) on this most challenging action validate that our hierarchical reward structure effectively decomposes complex manipulation into learnable sub-goals.

Table~\ref{tab:action_accuracy} presents the action classification accuracy of the Temporal Shift Module (TSM) with varying backbone depths. On the standard benchmark split (Test Set 1), our model achieves $89.97\%$ Top-1 accuracy with ResNet-101, demonstrating strong performance on fine-grained action recognition. On self-recorded clips (Test Set 2), the accuracy decreases to $71.11\%$, reflecting the domain shift between benchmark videos and real-world recordings. Notably, the Top-5 accuracy remains consistently high (above $94\%$) across both test sets, indicating that the correct action is typically among the model's top predictions even when the Top-1 prediction is incorrect. The performance gap between Test Set 1 and Test Set 2 (approximately $19\%$ for Top-1) highlights the challenge of generalizing from curated datasets to unconstrained real-world videos, motivating our object-centric approach.

Tables~\ref{tab:bleu_standard} and~\ref{tab:bleu_novel} compare video-to-command generation performance using BLEU scores. On standard object sets (Table~\ref{tab:bleu_standard}), our method with ResNet-101 backbone achieves BLEU-4 of 0.351, substantially outperforming the previous state-of-the-art Watch-and-Act (0.199 with InceptionV3). This represents a relative improvement of 76.4\% in BLEU-4 score. The ablation variants demonstrate the importance of each component: removing keyframe extraction reduces BLEU-4 to 0.298 ($-$15.1\%), while removing object selection reduces it to 0.237 ($-$32.5\%). On novel object sets (Table~\ref{tab:bleu_novel}), our method demonstrates strong zero-shot generalization, achieving BLEU-4 of 0.265 compared to 0.116 for Watch-and-Act—an improvement of 128.4\%. The smaller performance degradation between standard and novel sets (24.5\% vs. 41.7\% for Watch-and-Act) indicates that our Vision-Language Model integration effectively leverages semantic knowledge for unseen object categories. These results validate our hypothesis that decoupling action understanding from object recognition, combined with VLM-based object identification, yields superior generalization compared to end-to-end approaches.

\begin{table}[!ht]
\centering
\caption{Action classification accuracy (\%) of the Temporal Shift Module (TSM) with varying backbone depths. Test Set 1: standard benchmark split; Test Set 2: self-recorded clips for real-world evaluation. All models use 8-frame sampling with 8-clips $\times$ 3-crops augmentation using Modified Something-Something V2 during inference.}
\label{tab:action_accuracy}
\begin{tabularx}{1.0\textwidth} { 
   >{\centering\arraybackslash} m{0.2\textwidth} 
  || >{\centering\arraybackslash} m{0.2\textwidth} 
  | >{\centering\arraybackslash}m{0.25\textwidth}
  | >{\centering\arraybackslash} X }
\toprule
Test Set & Backbone & Top-1 Acc. & Top-5 Acc. \\
\midrule
\multirow{3}{*}{Test set 1}
& ResNet-34  & 82.79 & 93.50 \\
& ResNet-50  & 86.79 & 94.55 \\
& ResNet-101 & \textbf{89.97} & \textbf{96.82} \\
\midrule
\multirow{3}{*}{Test set 2}
& ResNet-34  & 63.70 & 94.81 \\
& ResNet-50  & 68.15 & 94.81 \\
& ResNet-101 & \textbf{71.11} & \textbf{95.56} \\
\bottomrule
\end{tabularx}
\end{table}

\begin{table}[!ht]
\centering
\caption{Video-to-command generation performance (BLEU scores) on standard object sets (\textbf{bold} and \underline{underline} are best and second best, respectively).}

\label{tab:bleu_standard}
\setlength{\tabcolsep}{6pt}
\begin{tabular}{llcccc}
\toprule
\textbf{Backbone} & \textbf{Method} & \textbf{BLEU-1} & \textbf{BLEU-2} & \textbf{BLEU-3} & \textbf{BLEU-4} \\
\midrule

\multirow{3}{*}{InceptionV3}
& Video2Command \cite{nguyen2018translating}        & 0.323 & 0.196 & 0.162 & 0.154 \\
& V2C \cite{nguyen2019v2cnet}                   & 0.355 & 0.227 & 0.201 & 0.164 \\
& Watch-and-Act \cite{yang2023watch}       & 0.399 & 0.287 & 0.263 & 0.199 \\

\midrule
\multirow{6}{*}{ResNet-50}
& Video2Command \cite{nguyen2018translating}        & 0.312 & 0.196 & 0.173 & 0.150 \\
& V2C \cite{nguyen2019v2cnet}                   & 0.357 & 0.231 & 0.201 & 0.153 \\
& Watch-and-Act \cite{yang2023watch}       & 0.394 & 0.271 & 0.248 & 0.187 \\
& Ours w/o key frame   & 0.562 & 0.464 & 0.361 & 0.297 \\
& Ours w/o object sel. & 0.521 & 0.407 & 0.300 & 0.237 \\
& Ours                 & \underline{0.607} & \underline{0.506} & \underline{0.397} & \underline{0.337} \\

\midrule
\multirow{6}{*}{ResNet-101}
& Video2Command \cite{nguyen2018translating}        & 0.319 & 0.183 & 0.175 & 0.148 \\
& V2C \cite{nguyen2019v2cnet}                   & 0.324 & 0.163 & 0.153 & 0.131 \\
& Watch-and-Act \cite{yang2023watch}       & 0.339 & 0.171 & 0.155 & 0.134 \\
& Ours w/o key frame   & 0.577 & 0.465 & 0.363 & 0.298 \\
& Ours w/o object sel. & 0.524 & 0.411 & 0.303 & 0.237 \\
& Ours                 & \textbf{0.618} & \textbf{0.511} & \textbf{0.403} & \textbf{0.351} \\
\bottomrule
\end{tabular}
\end{table}

\begin{table}[!ht]
\centering
\caption{Video-to-command generation performance (BLEU scores) on novel object sets not seen during training (\textbf{bold} and \underline{underline} are best and second best, respectively).}
\label{tab:bleu_novel}
\setlength{\tabcolsep}{6pt}
\begin{tabular}{llcccc}
\toprule
\textbf{Backbone} & \textbf{Method} & \textbf{BLEU-1} & \textbf{BLEU-2} & \textbf{BLEU-3} & \textbf{BLEU-4} \\
\midrule

\multirow{3}{*}{InceptionV3}
& Video2Command \cite{nguyen2018translating}  & 0.244 & 0.132 & 0.075 & 0.062 \\
& V2C \cite{nguyen2019v2cnet}             & 0.279 & 0.117 & 0.093 & 0.087 \\
& Watch-and-Act \cite{yang2023watch}  & 0.293 & 0.141 & 0.133 & 0.116 \\

\midrule
\multirow{6}{*}{ResNet-50}
& Video2Command \cite{nguyen2018translating}        & 0.305 & 0.150 & 0.090 & 0.082 \\
& V2C \cite{nguyen2019v2cnet}                   & 0.259 & 0.119 & 0.092 & 0.087 \\
& Watch-and-Act \cite{yang2023watch}        & 0.310 & 0.157 & 0.134 & 0.107 \\
& Ours w/o key frame   & 0.535 & 0.431 & 0.295 & 0.217 \\
& Ours w/o object sel. & 0.474 & 0.361 & 0.251 & 0.191 \\
& Ours                 & \textbf{0.577} & \underline{0.471} & \underline{0.356} & \underline{0.261}\\
\midrule
\multirow{6}{*}{ResNet-101}
& Video2Command \cite{nguyen2018translating}        & 0.226 & 0.107 & 0.094 & 0.081 \\
& V2C \cite{nguyen2019v2cnet}                   & 0.222 & 0.102 & 0.095 & 0.086 \\
& Watch-and-Act \cite{yang2023watch}        & 0.233 & 0.113 & 0.109 & 0.094 \\
& Ours w/o key frame   & 0.539 & 0.435 & 0.298 & 0.218 \\
& Ours w/o object sel. & 0.478 & 0.362 & 0.258 & 0.193 \\
& Ours                 & \underline{0.572} & \textbf{0.481} & \textbf{0.361} & \textbf{0.265} \\
\bottomrule
\end{tabular}
\end{table}

\subsubsection{Robot Manipulation Performance}

\begin{figure}[!ht]
    \centering
    \begin{overpic}[width=\textwidth, unit=1pt]{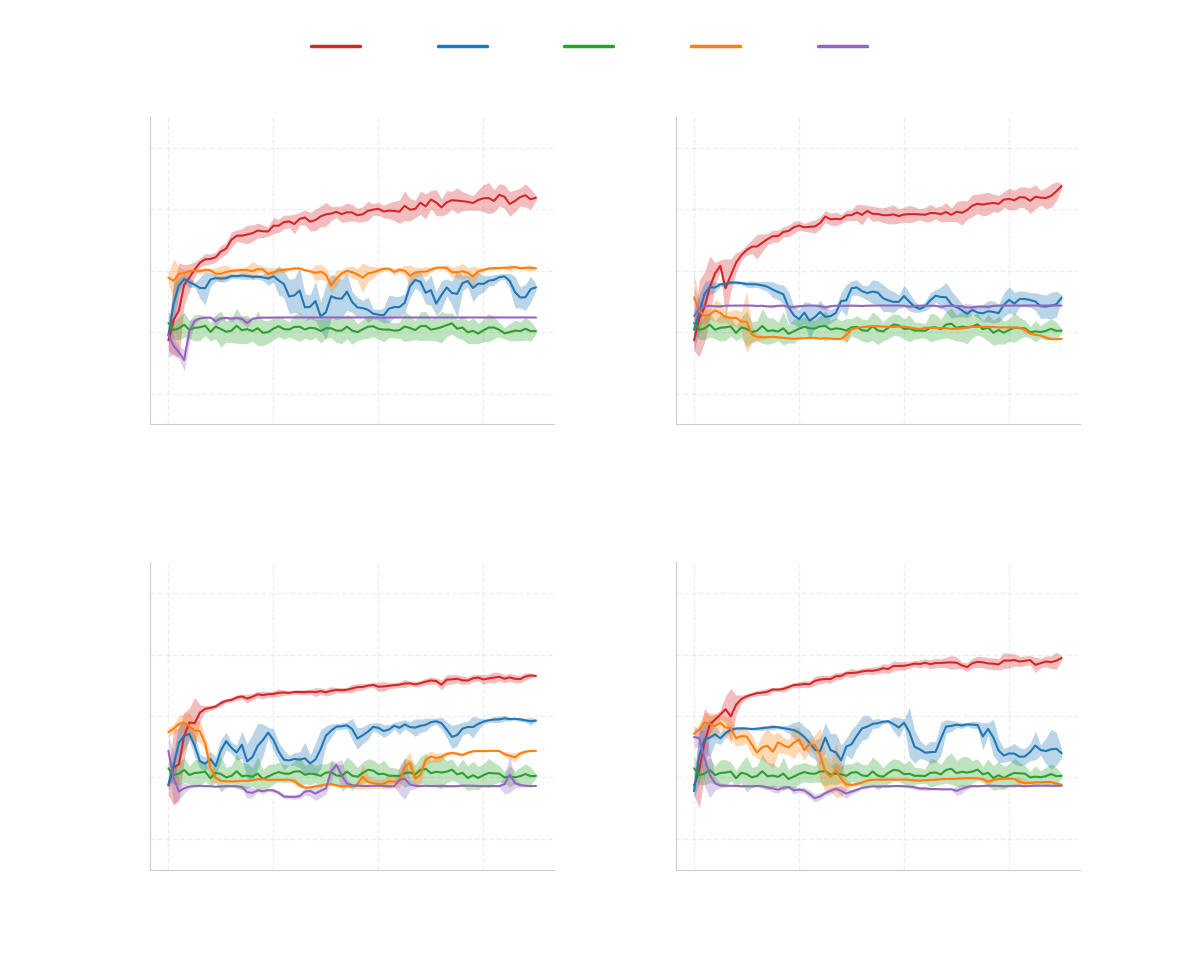}
        \put(13.6, 44.0){\footnotesize 0}
        \put(20.6, 44.0){\footnotesize 10000}
        \put(29.6, 44.0){\footnotesize 20000}
        \put(38.2, 44.0){\footnotesize 30000}
        \put(21.5, 41.0){\footnotesize Training episodes}
        \put(21.4, 38.0){\footnotesize (a) Reach action}
        \put(8.5, 48.0){\footnotesize -40}
        \put(8.5, 53.2){\footnotesize -20}
        \put(10.3, 58.1){\footnotesize 0}
        \put(9.3, 63.3){\footnotesize 20}
        \put(9.3, 68.4){\footnotesize 40}

        \put(13.6, 7.0){\footnotesize 0}
        \put(20.6, 7.0){\footnotesize 10000}
        \put(29.6, 7.0){\footnotesize 20000}
        \put(38.2, 7.0){\footnotesize 30000}
        \put(21.5, 4.0){\footnotesize Training episodes}
        \put(21.6, 1.0){\footnotesize (c) Move action}
        \put(8.5, 11.0){\footnotesize -40}
        \put(8.5, 16.2){\footnotesize -20}
        \put(10.3, 21.1){\footnotesize 0}
        \put(9.3, 27.3){\footnotesize 20}
        \put(9.3, 31.4){\footnotesize 40}

        \put(57.4, 44.0){\footnotesize 0}
        \put(64.4, 44.0){\footnotesize 10000}
        \put(73.4, 44.0){\footnotesize 20000}
        \put(82.0, 44.0){\footnotesize 30000}
        \put(65.3, 41.0){\footnotesize Training episodes}
        \put(65.9, 38.0){\footnotesize (b) Pick action}
        \put(52.3, 48.0){\footnotesize -40}
        \put(52.3, 53.2){\footnotesize -20}
        \put(54.1, 58.1){\footnotesize 0}
        \put(53.1, 63.3){\footnotesize 20}
        \put(53.1, 68.4){\footnotesize 40}

        \put(57.4, 7.0){\footnotesize 0}
        \put(64.4, 7.0){\footnotesize 10000}
        \put(73.4, 7.0){\footnotesize 20000}
        \put(82.0, 7.0){\footnotesize 30000}
        \put(65.3, 4.0){\footnotesize Training episodes}
        \put(66.2, 1.0){\footnotesize (d) Put action}
        \put(52.3, 11.0){\footnotesize -40}
        \put(52.3, 16.2){\footnotesize -20}
        \put(54.1, 21.1){\footnotesize 0}
        \put(53.1, 27.3){\footnotesize 20}
        \put(53.1, 31.4){\footnotesize 40}

        \put(5.1, 32.4){\footnotesize \rotatebox{90}{Average Reward}}
        \put(30.9, 77.2){\tiny TD3}
        \put(41.9, 77.2){\tiny SAC}
        \put(51.4, 77.2){\tiny DDPG}
        \put(62.3, 77.2){\tiny PPO}
        \put(73.7, 77.2){\tiny APPO}
    \end{overpic}
    \caption{ Comparison of DRL algorithm performance across four manipulation actions.}
    \label{fig:training_curves}
\end{figure}
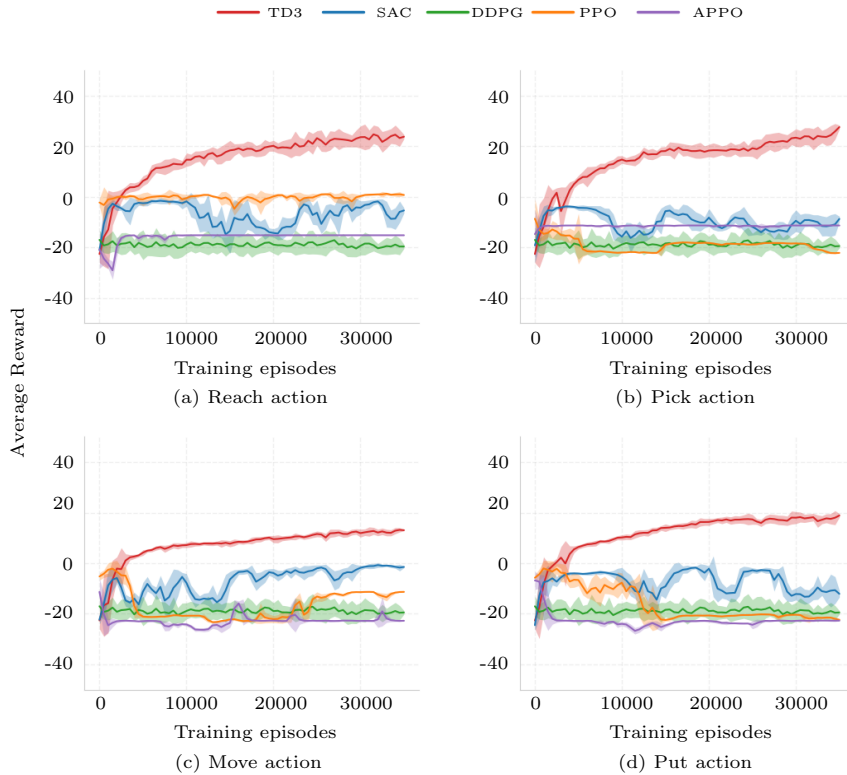

Table~\ref{tab:success_rate} reports task success rates under varying positional error thresholds. Our method achieves a $100\%$ success rate on the \textit{reach} action across all thresholds, demonstrating robust end-effector positioning. For more complex actions requiring object interaction, we observe strong performance: \textit{pick} achieves 70--100\% success depending on the threshold with an average of $87.5\%$, while \textit{move} and \textit{put} maintain 70--90\% success rates. The consistently high performance across thresholds from 1~cm to 4~cm indicates that our policy learns precise manipulation rather than relying on loose success criteria. Fig.~\ref{fig:training_curves} compares training convergence across five DRL algorithms: TD3 (ours), SAC, DDPG, PPO, and Asym-PPO. TD3 consistently achieves higher average returns and faster convergence across all four manipulation actions. For the \textit{reach} action, TD3 converges to positive returns within approximately 500 episodes, while PPO and Asym-PPO require over 1,500 episodes. The advantage of TD3 becomes more pronounced for complex actions: on \textit{put}, TD3 achieves stable positive returns around episode 2,000, whereas DDPG and SAC exhibit high variance throughout training.

\begin{figure*}[!ht]
    \centering
    \begin{overpic}[width=\textwidth, unit=1pt]{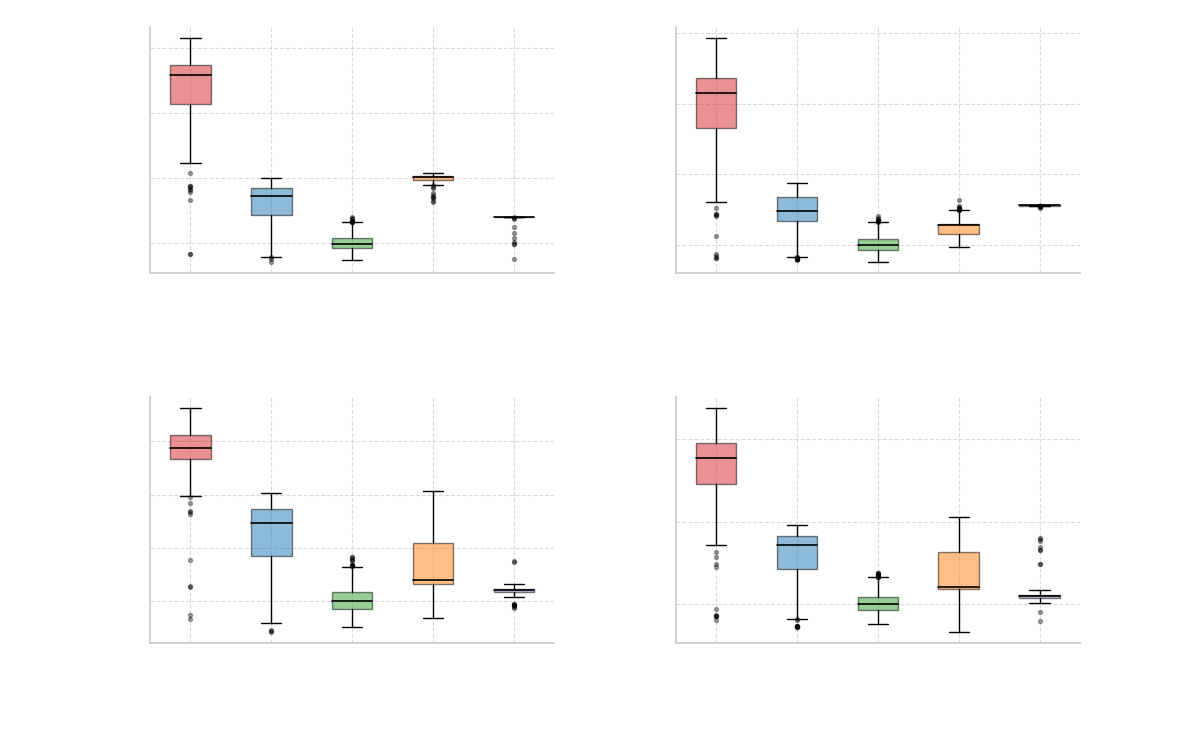}
        \put(14.0, 36.2){\footnotesize TD3}
        \put(20.6, 36.2){\footnotesize SAC}
        \put(26.8, 36.2){\footnotesize DDPG}
        \put(34.2, 36.2){\footnotesize PPO}
        \put(40.3, 36.2){\footnotesize APPO}
        \put(26.7, 33.7){\footnotesize Models}
        \put(21.4, 31.5){\footnotesize (a) Reach action}
        \put(8.5, 40.2){\footnotesize -0.2}
        \put(8.5, 45.4){\footnotesize -0.0}
        \put(9.2, 50.7){\footnotesize 0.2}
        \put(9.2, 56.2){\footnotesize 0.4}

        \put(58.0, 36.2){\footnotesize TD3}
        \put(64.6, 36.2){\footnotesize SAC}
        \put(70.8, 36.2){\footnotesize DDPG}
        \put(78.2, 36.2){\footnotesize PPO}
        \put(84.3, 36.2){\footnotesize APPO}
        \put(70.7, 33.7){\footnotesize Models}
        \put(65.6, 31.5){\footnotesize (b) Pick action}
        \put(52.5, 40.2){\footnotesize -0.2}
        \put(52.5, 45.4){\footnotesize -0.0}
        \put(53.2, 50.7){\footnotesize 0.2}
        \put(53.2, 56.2){\footnotesize 0.4}

        \put(14.0, 5.2){\footnotesize TD3}
        \put(20.6, 5.2){\footnotesize SAC}
        \put(26.8, 5.2){\footnotesize DDPG}
        \put(34.2, 5.2){\footnotesize PPO}
        \put(40.3, 5.2){\footnotesize APPO}
        \put(26.7, 2.7){\footnotesize Models}
        \put(21.4, 0.5){\footnotesize (c) Move action}
        \put(8.5, 9.2){\footnotesize -0.2}
        \put(8.5, 14.4){\footnotesize -0.0}
        \put(9.2, 19.2){\footnotesize 0.2}
        \put(9.2, 23.8){\footnotesize 0.4}

        \put(58.0, 5.2){\footnotesize TD3}
        \put(64.6, 5.2){\footnotesize SAC}
        \put(70.8, 5.2){\footnotesize DDPG}
        \put(78.2, 5.2){\footnotesize PPO}
        \put(84.3, 5.2){\footnotesize APPO}
        \put(70.7, 2.7){\footnotesize Models}
        \put(65.8, 0.5){\footnotesize (d) Put action}
        \put(52.5, 9.2){\footnotesize -0.2}
        \put(52.5, 14.4){\footnotesize -0.0}
        \put(53.2, 19.7){\footnotesize 0.2}
        \put(53.2, 23.8){\footnotesize 0.4}

        \put(5.1, 24.4){\footnotesize \rotatebox{90}{Average Reward}}
    \end{overpic}
    \caption{Box plots comparing average rewards for four manipulation actions. }
    \label{fig:boxplot_algorithms}
\end{figure*}

\begin{figure*}[t]
    \centering
    \begin{overpic}[width=\textwidth, unit=1pt]{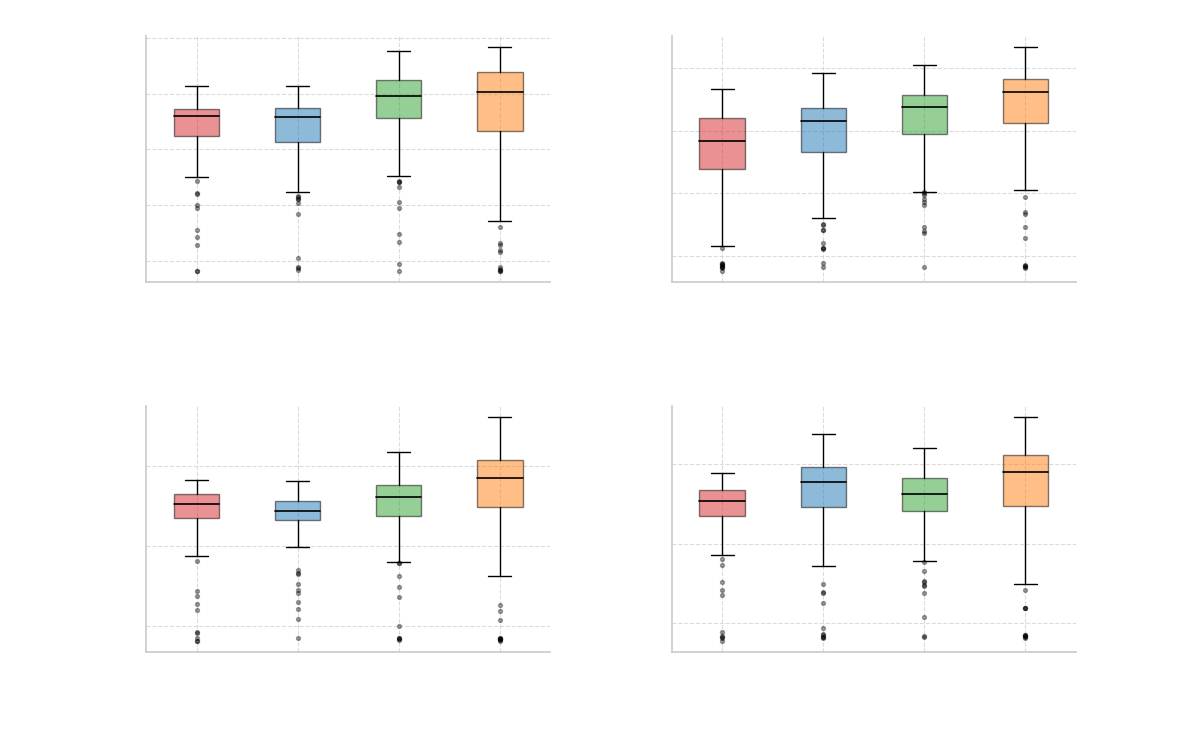}
        \put(14.5, 36.2){\footnotesize 1cm}
        \put(23.0, 36.2){\footnotesize 2cm}
        \put(31.8, 36.2){\footnotesize 3cm}
        \put(40.3, 36.2){\footnotesize 4cm}
        \put(22.7, 33.7){\footnotesize Posision error}
        \put(21.4, 31.5){\footnotesize (a) Reach action}
        \put(7.0, 39.7){\footnotesize -0.25}
        \put(7.0, 44.3){\footnotesize -0.00}
        \put(7.7, 48.8){\footnotesize 0.25}
        \put(7.7, 53.4){\footnotesize 0.50}

        \put(58.5, 36.2){\footnotesize 1cm}
        \put(67.0, 36.2){\footnotesize 2cm}
        \put(74.8, 36.2){\footnotesize 3cm}
        \put(84.3, 36.2){\footnotesize 4cm}
        \put(66.7, 33.7){\footnotesize Posision error}
        \put(65.9, 31.5){\footnotesize (b) Pick action}
        \put(51.0, 39.7){\footnotesize -0.25}
        \put(51.0, 44.3){\footnotesize -0.00}
        \put(51.7, 48.8){\footnotesize 0.25}
        \put(51.7, 53.4){\footnotesize 0.50}

        \put(14.5, 5.2){\footnotesize 1cm}
        \put(23.0, 5.2){\footnotesize 2cm}
        \put(31.8, 5.2){\footnotesize 3cm}
        \put(40.3, 5.2){\footnotesize 4cm}
        \put(22.7, 2.7){\footnotesize Posision error}
        \put(21.6, 0.5){\footnotesize (c) Move action}
        \put(7.0, 8.7){\footnotesize -0.25}
        \put(7.0, 13.3){\footnotesize -0.00}
        \put(7.7, 17.8){\footnotesize 0.25}
        \put(7.7, 22.4){\footnotesize 0.50}

        \put(58.5, 5.2){\footnotesize 1cm}
        \put(67.0, 5.2){\footnotesize 2cm}
        \put(74.8, 5.2){\footnotesize 3cm}
        \put(84.3, 5.2){\footnotesize 4cm}
        \put(66.7, 2.7){\footnotesize Posision error}
        \put(65.9, 0.5){\footnotesize (d) Put action}
        \put(51.0, 8.7){\footnotesize -0.25}
        \put(51.0, 13.3){\footnotesize -0.00}
        \put(51.7, 17.8){\footnotesize 0.25}
        \put(51.7, 22.4){\footnotesize 0.50}

        \put(5.1, 24.4){\footnotesize \rotatebox{90}{Average Reward}}
    \end{overpic}
    \caption{Box plots comparing average rewards under 4 different positional errors as the success threshold for four manipulation actions. }
    \label{fig:boxplot_precision}
\end{figure*}

Fig.~\ref{fig:boxplot_algorithms} presents box plots comparing final performance across algorithms. TD3 demonstrates superior median performance and lower variance compared to all baselines, particularly for \textit{pick}, \textit{move}, and \textit{put} actions. SAC achieves the second-best performance but exhibits notably higher variance, especially on \textit{move} actions. The on-policy methods (PPO, Asym-PPO) consistently underperform off-policy approaches, likely due to sample inefficiency in our highly randomized environment. Fig.~\ref{fig:boxplot_precision} analyzes the relationship between positional error threshold and average reward. As expected, relaxing the success threshold from 1~cm to 4~cm generally improves average rewards across all actions. However, our method maintains positive average returns even at the strictest 1~cm threshold for \textit{reach} and \textit{put} actions, demonstrating that the learned policies achieve sub-centimeter precision. The \textit{pick} and \textit{move} actions show more sensitivity to the threshold, reflecting the additional complexity of maintaining stable grasps during object manipulation.

\begin{table}[!ht]
\centering
\caption{Task success rates (\%) of our method under varying positional error thresholds.}
\label{tab:success_rate}
\begin{tabularx}{1.0\textwidth} { 
   >{\centering\arraybackslash} X
  || >{\centering\arraybackslash} m{0.18\textwidth} 
  | >{\centering \arraybackslash}m{0.18\textwidth}
  | >{\centering\arraybackslash} m{0.18\textwidth} 
  | >{\centering\arraybackslash} m{0.18\textwidth}  }
\toprule
\textbf{Action} & \textbf{1\,cm} & \textbf{2\,cm} & \textbf{3\,cm} & \textbf{4\,cm} \\
\midrule
Reach & 100 & 100 & 100 & 100 \\ \hline
Pick  & 70  & 80  & 100 & 90  \\ \hline
Move  & 90  & 80  & 70  & 90  \\ \hline
Put   & 90  & 90  & 70  & 90  \\
\bottomrule
\end{tabularx}
\end{table}

\subsubsection{Real-World Environment}
\begin{figure*}
    \centering
    \begin{subfigure}[!ht]{\textwidth}
    \centering
    \begin{overpic}[width=\textwidth, unit=1pt]{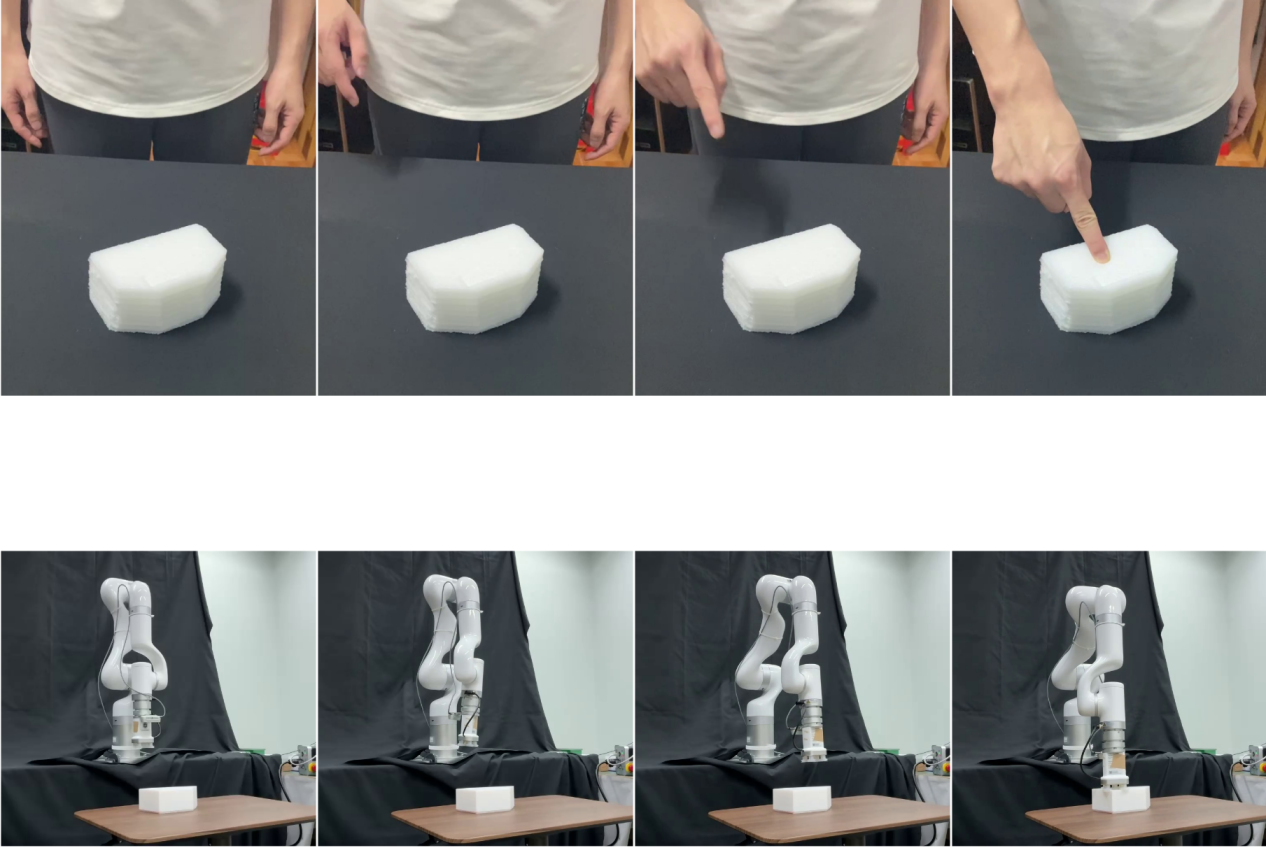}
        \put(49.2, 29.2){$\Bigg \downarrow$}
        \put(33.8, 29.){\texttt{``Touch the white trapezoid''}}
    \end{overpic}
    \caption{Reach action}
    \end{subfigure}
    \hfill
    \begin{subfigure}[!ht]{\textwidth}
    \centering
    \begin{overpic}[width=\textwidth, unit=1pt]{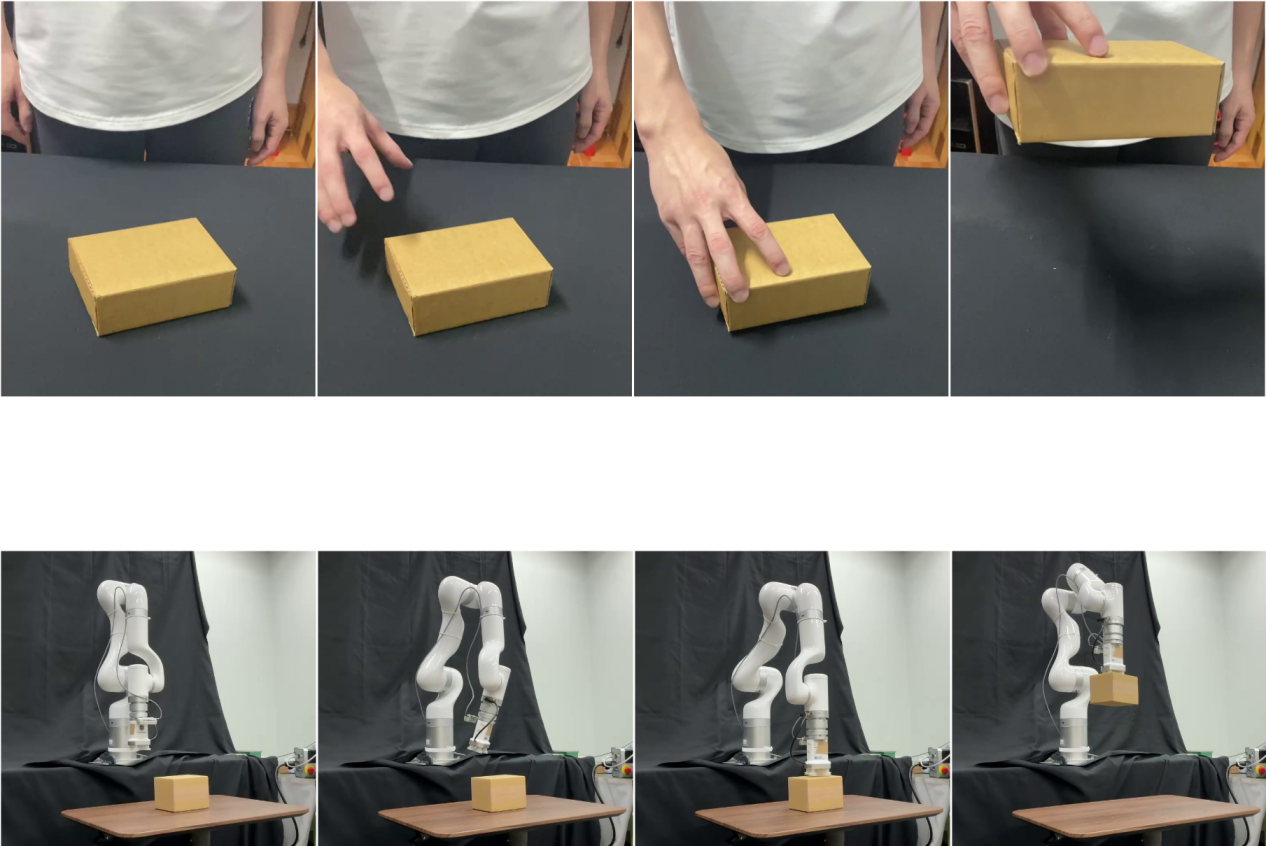}
        \put(49.2, 28.7){$\Bigg \downarrow$}
        \put(35.2, 28.7){\texttt{``Pick the brown box up''}}
    \end{overpic}
    \caption{Pick Action}
    \end{subfigure}
    \caption{Real experiment with UF850 manipulator}
    \label{fig:real_experiment}
\end{figure*}

To validate the practical applicability of our video understanding framework, we conducted preliminary experiments with a UF850 robotic manipulator in a real-world tabletop setting. The experimental setup consists of the 6-DoF UF850 arm equipped with a vacuum gripper, positioned in front of a wooden table with a manipulation object.
We evaluated two fundamental actions: \textit{reach} and \textit{pick}, as illustrated in Fig.~\ref{fig:real_experiment}. For each action, human demonstration videos were first collected in different environments and processed by our video understanding pipeline to generate manipulation commands. These commands, such as ``Touch the white trapezoid" and "Pick the brown box up, were then executed by the robot through a simple motion planning interface. In the reach task (Fig.~\ref{fig:real_experiment}a), the robot successfully navigates its end-effector to the target object position based on the generated command. In the pick task (Fig.~\ref{fig:real_experiment}b), the robot approaches the target object (a cardboard box), establishes contact with the vacuum gripper, and lifts it from the table surface. 

Table~\ref{tab:quantitative} summarizes the quantitative results over 10 trials per action. The system achieves success rates of 100\% for reach and 80\% for pick actions. Average execution times are 30s and 33s, respectively, with position accuracies of 25mm and 20mm. These results demonstrate that our framework effectively generalizes from training data to real-world scenarios with different backgrounds and lighting conditions.

\begin{table}[!ht]
\centering
\caption{Real-world experiments over 10 trials per action.}
\label{tab:quantitative}
\begin{tabularx}{1.0\textwidth} { 
   >{\centering\arraybackslash} X
  || >{\centering\arraybackslash} m{0.20\textwidth} 
  | >{\centering \arraybackslash}m{0.24\textwidth}
  | >{\centering\arraybackslash} m{0.28\textwidth}}
\toprule
\textbf{Action} & \textbf{Success rate} & \textbf{Execution time (s)} & \textbf{Position accuracy (mm)}\\
\midrule
Reach & 100 & 30 $\pm$ 3.8 & 25 $\pm$ 6\\ \hline
Pick  & 80  & 33 $\pm$ 5.4 & 20 $\pm$ 4\\
\bottomrule
\end{tabularx}
\end{table}


\section{Conclusion}
\label{sec:conclusion}
This paper presented a novel "Human-to-Robot" imitation learning framework that enables robots to acquire manipulation skills directly from unstructured video demonstrations. Our key contribution lies in the modular architecture that decouples the learning process into two specialized stages: Video Understanding for interpreting human demonstrations, and Robot Imitation for translating semantic commands into executable manipulation policies. The Video Understanding module integrates a Temporal Shift Module (TSM) for action classification with a Vision-Language Model (VLM) for object identification, achieving state-of-the-art performance on video-to-command generation with BLEU-4 scores of $0.351$ on standard objects and $0.265$ on novel objects, representing improvements of $76.4\%$ and $128.4\%$ over the best baseline, respectively. The Robot Imitation module employs TD3-based deep reinforcement learning with a carefully designed hierarchical reward structure, achieving an average success rate of $87.5\%$ across four fundamental manipulation actions: reach, pick, move, and put.
Our experimental results demonstrate three key advantages of the proposed approach. First, the decoupled architecture enables effective generalization to novel objects without requiring retraining, as the VLM leverages pretrained semantic knowledge. Second, the grammar-free command format provides a robust interface between video understanding and robot execution, eliminating ambiguities inherent in natural language descriptions. Third, the hierarchical reward design successfully decomposes complex manipulation into learnable sub-goals, enabling sub-centimeter precision even in cluttered environments with multiple distractors.

Despite these promising results, several limitations remain. The current framework is restricted to four fundamental manipulation primitives and single-arm operations, limiting applicability to more complex bimanual or tool-use tasks. The reliance on simulation for policy training introduces a sim-to-real gap that requires further investigation. Additionally, the object selection algorithm assumes clear visual separation between objects, which may fail in highly occluded scenarios. Several directions merit further exploration. Extending the action vocabulary to include more complex primitives such as pouring, stirring, and folding would broaden the framework's applicability. Incorporating tactile feedback and force control could improve manipulation robustness for deformable or fragile objects. Real-world deployment with domain adaptation techniques would validate the practical utility of our approach. Finally, integrating large language models (LLMs) for task planning could enable multi-step manipulation sequences from high-level natural language instructions.

\bibliography{ref}

@incollection{calinon2018learning,
  title={Learning from demonstration (programming by demonstration)},
  author={Calinon, Sylvain},
  booktitle={Encyclopedia of robotics},
  pages={1--8},
  year={2018},
  publisher={Springer}
}

@article{argall2009survey,
  title={A survey of robot learning from demonstration},
  author={Argall, Brenna D and Chernova, Sonia and Veloso, Manuela and Browning, Brett},
  journal={Robotics and autonomous systems},
  volume={57},
  number={5},
  pages={469--483},
  year={2009},
  publisher={Elsevier}
}

@inproceedings{karpathy2014large,
  title={Large-scale video classification with convolutional neural networks},
  author={Karpathy, Andrej and Toderici, George and Shetty, Sanketh and Leung, Thomas and Sukthankar, Rahul and Fei-Fei, Li},
  booktitle={Proceedings of the IEEE conference on Computer Vision and Pattern Recognition},
  pages={1725--1732},
  year={2014}
}

@inproceedings{akgun2012trajectories,
  title={Trajectories and keyframes for kinesthetic teaching: A human-robot interaction perspective},
  author={Akgun, Baris and Cakmak, Maya and Yoo, Jae Wook and Thomaz, Andrea Lockerd},
  booktitle={Proceedings of the seventh annual ACM/IEEE international conference on Human-Robot Interaction},
  pages={391--398},
  year={2012}
}

@inproceedings{sakr2020training,
  title={Training human teacher to improve robot learning from demonstration: A pilot study on kinesthetic teaching},
  author={Sakr, Maram and Freeman, Martin and Van der Loos, HF Machiel and Croft, Elizabeth},
  booktitle={2020 29th IEEE International Conference on Robot and Human Interactive Communication (RO-MAN)},
  pages={800--806},
  year={2020},
  organization={IEEE}
}

@article{yu2018one,
  title={One-shot imitation from observing humans via domain-adaptive meta-learning},
  author={Yu, Tianhe and Finn, Chelsea and Xie, Annie and Dasari, Sudeep and Zhang, Tianhao and Abbeel, Pieter and Levine, Sergey},
  journal={arXiv preprint arXiv:1802.01557},
  year={2018}
}

@inproceedings{maccio2024kinesthetic,
  title={Kinesthetic Teaching in Robotics: a Mixed Reality Approach},
  author={Macci{\`o}, Simone and Shaaban, Mohamad and Carf{\`\i}, Alessandro and Mastrogiovanni, Fulvio},
  booktitle={2024 33rd IEEE International Conference on Robot and Human Interactive Communication (ROMAN)},
  pages={1610--1617},
  year={2024},
  organization={IEEE}
}

@inproceedings{gomez2012kinesthetic,
  title={Kinesthetic teaching via fast marching square},
  author={G{\'o}mez, Javier V and Alvarez, David and Garrido, Santiago and Moreno, Luis},
  booktitle={2012 IEEE/RSJ International Conference on Intelligent Robots and Systems},
  pages={1305--1310},
  year={2012},
  organization={IEEE}
}

@inproceedings{fu2023teleoperation,
  title={Teleoperation Method For Controlling Robotic Arm Based On Multi-channel EMG Signals},
  author={Fu, Rongrong and Li, Qisen and Wang, Shiwei and Sun, Guangbin},
  booktitle={2023 38th Youth Academic Annual Conference of Chinese Association of Automation (YAC)},
  pages={1264--1268},
  year={2023},
  organization={IEEE}
}

@inproceedings{peters2016feasibility,
  title={Feasibility of physiotherapy exercise capturing using a low-cost motion capture system},
  author={Peters, Konrad and Unterbrunner, Alexander and Martinek, Daniel and Hofmann, Alexander and Hlavacs, Helmut},
  booktitle={International Conference on Computer Games, Multimedia \& Allied Technology (CGAT). Proceedings},
  pages={47},
  year={2016},
  organization={Global Science and Technology Forum}
}

@inproceedings{koenemann2014real,
  title={Real-time imitation of human whole-body motions by humanoids},
  author={Koenemann, Jonas and Burget, Felix and Bennewitz, Maren},
  booktitle={2014 IEEE International Conference on Robotics and Automation (ICRA)},
  pages={2806--2812},
  year={2014},
  organization={IEEE}
}

@article{racinskis2022motion,
  title={A motion capture and imitation learning based approach to robot control},
  author={Racinskis, Peteris and Arents, Janis and Greitans, Modris},
  journal={Applied Sciences},
  volume={12},
  number={14},
  pages={7186},
  year={2022},
  publisher={MDPI}
}

@article{yang2016teleoperation,
  title={Teleoperation control based on combination of wave variable and neural networks},
  author={Yang, Chenguang and Wang, Xingjian and Li, Zhijun and Li, Yanan and Su, Chun-Yi},
  journal={IEEE Transactions on Systems, Man, and Cybernetics: Systems},
  volume={47},
  number={8},
  pages={2125--2136},
  year={2016},
  publisher={IEEE}
}

@article{smith2019avid,
  title={Avid: Learning multi-stage tasks via pixel-level translation of human videos},
  author={Smith, Laura and Dhawan, Nikita and Zhang, Marvin and Abbeel, Pieter and Levine, Sergey},
  journal={arXiv preprint arXiv:1912.04443},
  year={2019}
}

@inproceedings{nguyen2018translating,
  title={Translating videos to commands for robotic manipulation with deep recurrent neural networks},
  author={Nguyen, Anh and Kanoulas, Dimitrios and Muratore, Luca and Caldwell, Darwin G and Tsagarakis, Nikos G},
  booktitle={2018 IEEE International Conference on Robotics and Automation (ICRA)},
  pages={3782--3788},
  year={2018},
  organization={IEEE}
}

@article{nguyen2019v2cnet,
  title={V2cnet: A deep learning framework to translate videos to commands for robotic manipulation},
  author={Nguyen, Anh and Do, Thanh-Toan and Reid, Ian and Caldwell, Darwin G and Tsagarakis, Nikos G},
  journal={arXiv preprint arXiv:1903.10869},
  year={2019}
}

@article{zhang2019reconstruct,
  title={Reconstruct and represent video contents for captioning via reinforcement learning},
  author={Zhang, Wei and Wang, Bairui and Ma, Lin and Liu, Wei},
  journal={IEEE transactions on pattern analysis and machine intelligence},
  volume={42},
  number={12},
  pages={3088--3101},
  year={2019},
  publisher={IEEE}
}

@article{yang2023watch,
  title={Watch and act: Learning robotic manipulation from visual demonstration},
  author={Yang, Shuo and Zhang, Wei and Song, Ran and Cheng, Jiyu and Wang, Hesheng and Li, Yibin},
  journal={IEEE Transactions on Systems, Man, and Cybernetics: Systems},
  volume={53},
  number={7},
  pages={4404--4416},
  year={2023},
  publisher={IEEE}
}

@inproceedings{do2018affordancenet,
  title={Affordancenet: An end-to-end deep learning approach for object affordance detection},
  author={Do, Thanh-Toan and Nguyen, Anh and Reid, Ian},
  booktitle={2018 IEEE international conference on robotics and automation (ICRA)},
  pages={5882--5889},
  year={2018},
  organization={IEEE}
}

@inproceedings{shan2020understanding,
  title={Understanding human hands in contact at internet scale},
  author={Shan, Dandan and Geng, Jiaqi and Shu, Michelle and Fouhey, David F},
  booktitle={Proceedings of the IEEE/CVF conference on computer vision and pattern recognition},
  pages={9869--9878},
  year={2020}
}

@inproceedings{lin2019tsm,
  title={Tsm: Temporal shift module for efficient video understanding},
  author={Lin, Ji and Gan, Chuang and Han, Song},
  booktitle={Proceedings of the IEEE/CVF international conference on computer vision},
  pages={7083--7093},
  year={2019}
}

@inproceedings{venugopalan2015sequence,
  title={Sequence to sequence-video to text},
  author={Venugopalan, Subhashini and Rohrbach, Marcus and Donahue, Jeffrey and Mooney, Raymond and Darrell, Trevor and Saenko, Kate},
  booktitle={Proceedings of the IEEE international conference on computer vision},
  pages={4534--4542},
  year={2015}
}

@inproceedings{zhang2020grasp,
  title={Grasp for stacking via deep reinforcement learning},
  author={Zhang, Junhao and Zhang, Wei and Song, Ran and Ma, Lin and Li, Yibin},
  booktitle={2020 IEEE International Conference on Robotics and Automation (ICRA)},
  pages={2543--2549},
  year={2020},
  organization={IEEE}
}

@inproceedings{kalashnikov2018scalable,
  title={Scalable deep reinforcement learning for vision-based robotic manipulation},
  author={Kalashnikov, Dmitry and Irpan, Alex and Pastor, Peter and Ibarz, Julian and Herzog, Alexander and Jang, Eric and Quillen, Deirdre and Holly, Ethan and Kalakrishnan, Mrinal and Vanhoucke, Vincent and others},
  booktitle={Conference on robot learning},
  pages={651--673},
  year={2018},
  organization={PMLR}
}

@inproceedings{zeng2018learning,
  title={Learning synergies between pushing and grasping with self-supervised deep reinforcement learning},
  author={Zeng, Andy and Song, Shuran and Welker, Stefan and Lee, Johnny and Rodriguez, Alberto and Funkhouser, Thomas},
  booktitle={2018 IEEE/RSJ International Conference on Intelligent Robots and Systems (IROS)},
  pages={4238--4245},
  year={2018},
  organization={IEEE}
}

@article{zeng2022robotic,
  title={Robotic pick-and-place of novel objects in clutter with multi-affordance grasping and cross-domain image matching},
  author={Zeng, Andy and Song, Shuran and Yu, Kuan-Ting and Donlon, Elliott and Hogan, Francois R and Bauza, Maria and Ma, Daolin and Taylor, Orion and Liu, Melody and Romo, Eudald and others},
  journal={The International Journal of Robotics Research},
  volume={41},
  number={7},
  pages={690--705},
  year={2022},
  publisher={SAGE Publications Sage UK: London, England}
}

@article{jang2017end,
  title={End-to-end learning of semantic grasping},
  author={Jang, Eric and Vijayanarasimhan, Sudheendra and Pastor, Peter and Ibarz, Julian and Levine, Sergey},
  journal={arXiv preprint arXiv:1707.01932},
  year={2017}
}

@inproceedings{shridhar2022cliport,
  title={Cliport: What and where pathways for robotic manipulation},
  author={Shridhar, Mohit and Manuelli, Lucas and Fox, Dieter},
  booktitle={Conference on robot learning},
  pages={894--906},
  year={2022},
  organization={PMLR}
}

@article{liu2023visual,
  title={Visual instruction tuning},
  author={Liu, Haotian and Li, Chunyuan and Wu, Qingyang and Lee, Yong Jae},
  journal={Advances in neural information processing systems},
  volume={36},
  pages={34892--34916},
  year={2023}
}

@inproceedings{bansal2016blur,
  title={Blur image detection using Laplacian operator and Open-CV},
  author={Bansal, Raghav and Raj, Gaurav and Choudhury, Tanupriya},
  booktitle={2016 International Conference System Modeling \& Advancement in Research Trends (SMART)},
  pages={63--67},
  year={2016},
  organization={IEEE}
}

@misc{coumans2016pybullet,
  author = {Erwin Coumans and Yunfei Bai},
  title = {PyBullet, a Python module for physics simulation for games, robotics and machine learning},
  year = {2016}
}

@inproceedings{zitkovich2023rt,
  title={Rt-2: Vision-language-action models transfer web knowledge to robotic control},
  author={Zitkovich, Brianna and Yu, Tianhe and Xu, Sichun and Xu, Peng and Xiao, Ted and Xia, Fei and Wu, Jialin and Wohlhart, Paul and Welker, Stefan and Wahid, Ayzaan and others},
  booktitle={Conference on Robot Learning},
  pages={2165--2183},
  year={2023},
  organization={PMLR}
}

@article{driess2023palm,
  title={Palm-e: An embodied multimodal language model},
  author={Driess, Danny and Xia, Fei and Sajjadi, Mehdi SM and Lynch, Corey and Chowdhery, Aakanksha and Wahid, Ayzaan and Tompson, Jonathan and Vuong, Quan and Yu, Tianhe and Huang, Wenlong and others},
  year={2023}
}

@article{zeng2022socratic,
  title={Socratic models: Composing zero-shot multimodal reasoning with language},
  author={Zeng, Andy and Attarian, Maria and Ichter, Brian and Choromanski, Krzysztof and Wong, Adrian and Welker, Stefan and Tombari, Federico and Purohit, Aveek and Ryoo, Michael and Sindhwani, Vikas and others},
  journal={arXiv preprint arXiv:2204.00598},
  year={2022}
}

@article{huang2022inner,
  title={Inner monologue: Embodied reasoning through planning with language models},
  author={Huang, Wenlong and Xia, Fei and Xiao, Ted and Chan, Harris and Liang, Jacky and Florence, Pete and Zeng, Andy and Tompson, Jonathan and Mordatch, Igor and Chebotar, Yevgen and others},
  journal={arXiv preprint arXiv:2207.05608},
  year={2022}
}

@inproceedings{fujimoto2018addressing,
  title={Addressing function approximation error in actor-critic methods},
  author={Fujimoto, Scott and Hoof, Herke and Meger, David},
  booktitle={International conference on machine learning},
  pages={1587--1596},
  year={2018},
  organization={PMLR}
}

@article{correia2024survey,
  title={A survey of demonstration learning},
  author={Correia, Andre and Alexandre, Luis A},
  journal={Robotics and Autonomous Systems},
  volume={182},
  pages={104812},
  year={2024},
  publisher={Elsevier}
}

@article{hwang2018real,
  title={Real-time pose imitation by mid-size humanoid robot with servo-cradle-head RGB-D vision system},
  author={Hwang, Chih-Lyang and Liao, Guo-Hsuan},
  journal={IEEE Transactions on Systems, Man, and Cybernetics: Systems},
  volume={49},
  number={1},
  pages={181--191},
  year={2018},
  publisher={IEEE}
}

@inproceedings{liu2018imitation,
  title={Imitation from observation: Learning to imitate behaviors from raw video via context translation},
  author={Liu, YuXuan and Gupta, Abhishek and Abbeel, Pieter and Levine, Sergey},
  booktitle={2018 IEEE international conference on robotics and automation (ICRA)},
  pages={1118--1125},
  year={2018},
  organization={IEEE}
}

@inproceedings{zhu2017unpaired,
  title={Unpaired image-to-image translation using cycle-consistent adversarial networks},
  author={Zhu, Jun-Yan and Park, Taesung and Isola, Phillip and Efros, Alexei A},
  booktitle={Proceedings of the IEEE international conference on computer vision},
  pages={2223--2232},
  year={2017}
}

@article{jin2024robotgpt,
  title={Robotgpt: Robot manipulation learning from chatgpt},
  author={Jin, Yixiang and Li, Dingzhe and Shi, Jun and Hao, Peng and Sun, Fuchun and Zhang, Jianwei and Fang, Bin and others},
  journal={IEEE Robotics and Automation Letters},
  volume={9},
  number={3},
  pages={2543--2550},
  year={2024},
  publisher={IEEE}
}

@article{zhou2024visual,
  title={Visual in-context learning for large vision-language models},
  author={Zhou, Yucheng and Li, Xiang and Wang, Qianning and Shen, Jianbing},
  journal={arXiv preprint arXiv:2402.11574},
  year={2024}
}

@inproceedings{deng2009imagenet,
  title={Imagenet: A large-scale hierarchical image database},
  author={Deng, Jia and Dong, Wei and Socher, Richard and Li, Li-Jia and Li, Kai and Fei-Fei, Li},
  booktitle={2009 IEEE conference on computer vision and pattern recognition},
  pages={248--255},
  year={2009},
  organization={Ieee}
}

@article{zhai2021one,
  title={One-shot object affordance detection in the wild},
  author={Zhai, Wei and Luo, Hongchen and Zhang, Jing and Cao, Yang and Tao, Dacheng},
  journal={arXiv preprint arXiv:2108.03658},
  year={2021}
}

@article{mao2025robot,
  title={Robot Learning from a Physical World Model},
  author={Mao, Jiageng and He, Sicheng and Wu, Hao-Ning and You, Yang and Sun, Shuyang and Wang, Zhicheng and Bao, Yanan and Chen, Huizhong and Guibas, Leonidas and Guizilini, Vitor and others},
  journal={arXiv preprint arXiv:2511.07416},
  year={2025}
}

@article{lenz2015deep,
  title={Deep learning for detecting robotic grasps},
  author={Lenz, Ian and Lee, Honglak and Saxena, Ashutosh},
  journal={The International Journal of Robotics Research},
  volume={34},
  number={4-5},
  pages={705--724},
  year={2015},
  publisher={SAGE Publications Sage UK: London, England}
}

@article{eze2025learning,
  title={Learning by watching: A review of video-based learning approaches for robot manipulation},
  author={Eze, Chrisantus and Crick, Christopher},
  journal={IEEE Access},
  year={2025},
  publisher={IEEE}
}

@article{qu2025spatialvla,
  title={Spatialvla: Exploring spatial representations for visual-language-action model},
  author={Qu, Delin and Song, Haoming and Chen, Qizhi and Yao, Yuanqi and Ye, Xinyi and Ding, Yan and Wang, Zhigang and Gu, JiaYuan and Zhao, Bin and Wang, Dong and others},
  journal={arXiv preprint arXiv:2501.15830},
  year={2025}
}

@inproceedings{zhao2025taste,
  title={TASTE-Rob: Advancing video generation of task-oriented hand-object interaction for generalizable robotic manipulation},
  author={Zhao, Hongxiang and Liu, Xingchen and Xu, Mutian and Hao, Yiming and Chen, Weikai and Han, Xiaoguang},
  booktitle={Proceedings of the Computer Vision and Pattern Recognition Conference},
  pages={27683--27693},
  year={2025}
}

@inproceedings{he2016deep,
  title={Deep residual learning for image recognition},
  author={He, Kaiming and Zhang, Xiangyu and Ren, Shaoqing and Sun, Jian},
  booktitle={Proceedings of the IEEE conference on computer vision and pattern recognition},
  pages={770--778},
  year={2016}
}

@inproceedings{haarnoja2018soft,
  title={Soft actor-critic: Off-policy maximum entropy deep reinforcement learning with a stochastic actor},
  author={Haarnoja, Tuomas and Zhou, Aurick and Abbeel, Pieter and Levine, Sergey},
  booktitle={International conference on machine learning},
  pages={1861--1870},
  year={2018},
  organization={Pmlr}
}

@article{lillicrap2015continuous,
  title={Continuous control with deep reinforcement learning},
  author={Lillicrap, Timothy P and Hunt, Jonathan J and Pritzel, Alexander and Heess, Nicolas and Erez, Tom and Tassa, Yuval and Silver, David and Wierstra, Daan},
  journal={arXiv preprint arXiv:1509.02971},
  year={2015}
}

@article{schulman2017proximal,
  title={Proximal policy optimization algorithms},
  author={Schulman, John and Wolski, Filip and Dhariwal, Prafulla and Radford, Alec and Klimov, Oleg},
  journal={arXiv preprint arXiv:1707.06347},
  year={2017}
}

@article{jeon2023learning,
  title={Learning whole-body manipulation for quadrupedal robot},
  author={Jeon, Seunghun and Jung, Moonkyu and Choi, Suyoung and Kim, Beomjoon and Hwangbo, Jemin},
  journal={IEEE Robotics and Automation Letters},
  volume={9},
  number={1},
  pages={699--706},
  year={2023},
  publisher={IEEE}
}

\end{document}